\documentclass[acmtog]{acmart}

\acmSubmissionID{499}   

\setcitestyle{square}

\usepackage{soul}
\soulregister\ref7
\soulregister\cite7
\soulregister\citeetal7
\soulregister\refSec7
\soulregister\refFig7
\soulregister\refTbl7
\soulregister\refEq7
\soulregister\cite7
\soulregister\ref7
\soulregister\pageref7
\soulregister\shortcite7
\soulregister\eg7
\soulregister\ie7
\soulregister\etal7
\soulregister\unsure7
\soulregister\todo7
\soulregister\smoothingkernel7
\soulregister\offset7

\usepackage{microtype}
\usepackage{multicol}
\usepackage{amsmath}
\usepackage{amsfonts}
\usepackage{booktabs}
\usepackage{hyperref}
\usepackage{multirow}
\usepackage{graphicx}
\usepackage{dsfont}
\usepackage{pifont}
\usepackage{xcolor}
\usepackage{wrapfig}
\usepackage{nicefrac}
\usepackage{flushend}
\usepackage{ctable}
\usepackage{natbib}
\usepackage{xspace}
\usepackage{algorithm}
\usepackage{algpseudocode}
\usepackage{comment}
\usepackage{soul}

\usepackage[normalem]{ulem}
\usepackage{caption}
\usepackage{subcaption}

\DeclareGraphicsExtensions{.png,.jpg,.pdf,.ai,.psd}
\graphicspath{{images/}}

\def\myfigure#1#2{%
    \begin{figure}[tb]%
    \centering\includegraphics*[width = \linewidth]{#1}%
    \vspace{-.3cm}%
    \caption{#2}%
    \label{fig:#1}%
    \end{figure}%
}

\def\mycfigure#1#2{%
    \begin{figure*}[htb]%
    \centering\includegraphics*[width = \linewidth]{#1}%
    \vspace{-.2cm}%
    \caption{#2}%
    \label{fig:#1}%
    \end{figure*}%
}

\newcommand{\mysection}[2]{\section{#1}\label{sec:#2}}
\newcommand{\mysubsection}[2]{\subsection{#1}\label{sec:#2}}
\newcommand{\mysubsubsection}[2]{\subsubsection{#1}\label{sec:#2}}

\newcommand{\refSec}[1]{Sec.~\ref{sec:#1}}
\newcommand{\refFig}[1]{Fig.~\ref{fig:#1}}
\newcommand{\refEq}[1]{Eq.~\ref{eq:#1}}
\newcommand{\refTab}[1]{Tab.~\ref{tab:#1}}
\newcommand{\refAlg}[1]{Alg.~\ref{alg:#1}}

\newcommand{\ie}{i.e.,\ }
\newcommand{\eg}{e.g.,\ }

\newcommand{\mymath}[2]{\newcommand{#1}{\TextOrMath{$#2$\xspace}{#2}}}

\newcommand{\unsure}[1]{{\sethlcolor{yellow}\hl{#1}}}
\newcommand{\todo}[1]{{\sethlcolor{green}\hl{#1}}}

\definecolor{colorNoSmooth}{HTML}{4285f4}      
\definecolor{colorNoNN}{HTML}{ea4335}      
\definecolor{colorNoLocal}{HTML}{ff9900}      
\definecolor{colorFD}{HTML}{34a853}      
\definecolor{colorFR22}{HTML}{ff00ff}
\definecolor{colorsFast}{HTML}{46bdc6}
\definecolor{colorOurs}{HTML}{202020}

\definecolor{colorLHS}{HTML}{000000}
\definecolor{colorGaussian}{HTML}{000000}
\definecolor{colorAdaptive}{HTML}{000000}
\definecolor{colorMASA}{HTML}{000000}
\definecolor{grey}{HTML}{282828}

\newcommand{\task}[1]{\textsc{\textsf{\textbf{#1}}}}
\newcommand{\method}[1]{\textcolor{color#1}{\texttt{\textbf{#1}}}}
\newcommand{\almostMethod}[1]{\textcolor{grey}{\texttt{\textbf{#1}}}}

\sethlcolor{neonGreen}

\newcommand{\changed}[1]{\textcolor{black}{#1}}

\NewDocumentCommand{\rot}{O{45} O{1em}m}{\makebox[#2][l]{\rotatebox{#1}{#3}}}%

\usepackage{siunitx}
\usepackage{caption}

\usepackage{pifont}
\newcommand{\cmark}{\checkmark}%
\newcommand{\xmark}{\scalebox{0.85}{\ding{53}}}

\usepackage{xr}
\makeatletter
\newcommand*{\addFileDependency}[1]{
  \typeout{(#1)}
  \@addtofilelist{#1}
  \IfFileExists{#1}{}{\typeout{No file #1.}}
}
\makeatother

\newcommand*{\myexternaldocument}[1]{%
    \externaldocument[supplemental-]{#1}%
    \addFileDependency{#1.tex}%
    \addFileDependency{#1.aux}%
}

\usepackage{icomma}

\myexternaldocument{supplemental}

\usepackage[nolist]{acronym}
\begin{acronym}
\acro{AD}{automatic differentiation}
\acro{BN}{batch normalization}
\acro{BRDF}{bi-directional reflectance distribution function}
\acro{BTF}{bi-directional texture function}
\acro{CNN}{convolutional neural network}
\acro{DA}{dual annealing}
\acro{DGS}{directional Gaussian smoothing}
\acro{DSSIM}{structural dissimilarity}
\acro{FD}{finite differences}
\acro{GA}{genetic algorithm}
\acro{GD}{gradient descent}
\acro{IOR}{index of refraction}
\acro{KL}{Kullback-Leibler}
\acro{KLD}{Kullback-Leibler divergence}
\acro{LHS}{Latin hypercube sampling}
\acro{MAE}{mean absolute error}
\acro{MASA}{mixed adaptive sampling algorithm}
\acro{MC}{Monte Carlo}
\acro{MCMC}{Markov-chain Monte Carlo}
\acro{ML}{machine learning}
\acro{MLP}{multi-layered perceptron}
\acro{MSE}{mean-squared error}
\acro{NAS}{neural architecture search}
\acro{NeRF}{neural radiance fields}
\acro{NN}{neural network}
\acro{NLL}{negative $\log$-likelihood}
\acro{NP}{neural proxy}
\acro{ODE}{ordinary differential equation}
\acro{PCA}{principal component analysis}
\acro{QMC}{quasi-Monte Carlo}
\acro{PDF}{probability density function}
\acro{RBF}{radial basis function}
\acro{RSM}{response surface map}
\acro{SA}{simulated annealing}
\acro{SPSA}{simultaneous perturbation stochastic approximation}
\acro{SDF}{signed distance function}
\acro{SSIM}{structural similarity}
\acro{svBRDF}{spatially-varying BRDF}
\acro{VAE}{variational autoencoder}

\end{acronym}

\mymath{\adjoint}a
\mymath{\lossOffset}{\rho}
\mymath{\learningRate}{\alpha}
\mymath{\locality}{\lambda}
\mymath{\mcmcSample}{y}
\mymath{\mcmcSampleGradient}{\mathbf z}
\mymath{\mcmcState}{\bar}
\mymath{\objective}f
\mymath{\offset}{\tau}
\mymath{\parameters}{\theta}
\mymath{\parameterDimension}n
\mymath{\parameterSpace}{\Theta}
\mymath{\sampleCount}{N}
\mymath{\smoothObjective}g
\mymath{\smoothingKernel}{\kappa}
\mymath{\spreadinner}{\sigma_i}
\mymath{\spreadouter}{\sigma_o}
\mymath{\surrogate}h
\mymath{\surrogateLoss}l
\mymath{\surrogateParameters}{\phi}
\mymath{\surrogateParameterSpace}{\Phi}
\mymath{\surrogateParameterDimension}m
\mymath{\potentialMatrix}{\textsf{M}}
\mymath{\inputvector}{\textrm{x}}

\mymath{\nsplinepts}{10}
\mymath{\numtexels}{196,608}
\mymath{\texturedim}{256}
\mymath{\numvertices}{2,562}
\mymath{\meshdim}{7,686}
\mymath{\numweights}{35,152}
\mymath{\numsplineweights}{8,764}

\def\myparagraph#1{\paragraph{#1}}

\MakeRobust{\Call}

\newcommand{\name}{ZeroGrads\xspace}

\setcopyright{rightsretained}
\acmJournal{TOG}
\acmYear{2024}
\acmVolume{43} 
\acmNumber{4} 
\acmArticle{49}
\acmMonth{7}
\acmDOI{10.1145/3658173}

\ccsdesc[500]{Computing methodologies~Rendering}

\begin{document}

\title[ZeroGrads]{ZeroGrads: Learning Local Surrogates for Non-Differentiable Graphics}

\author{Michael Fischer}
\affiliation{%
	\institution{University College London}
	\country{United Kingdom}
}
\email{m.fischer@cs.ucl.ac.uk}

\author{Tobias Ritschel}
\affiliation{%
	\institution{University College London}
	\country{United Kingdom}
}
\email{t.ritschel@ucl.ac.uk}

\keywords{Differentiable Rendering; Inverse Rendering; Surrogate Learning; Gradient Descent; Optimization}

\begin{teaserfigure}
    \includegraphics[width=\textwidth]{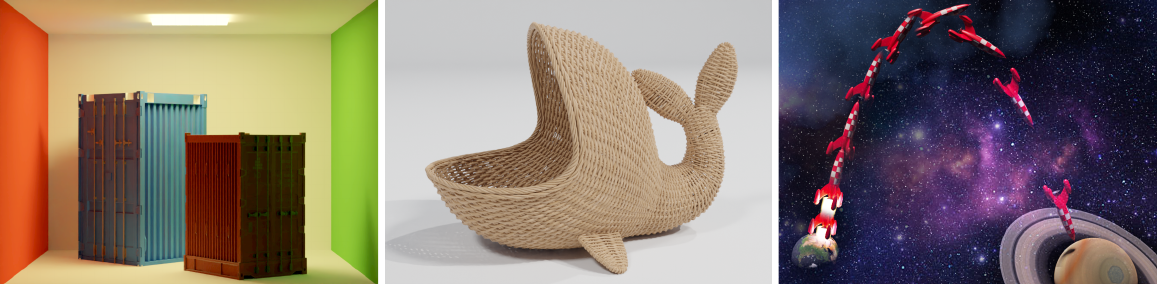}
    \caption{
    Our method optimizes arbitrary (black-box) graphics pipelines, which all do not trivially provide gradients, such as rendering (left: finding the box and light position under a discontinuous integrand),  modelling (middle: discrete number of vertical and horizontal wicker stakes), and animation (right: optimal control of the engine turn-off time), by fitting a neural network to the loss landscape and then using the network's gradients for parameter optimization.
    }
    \label{fig:Teaser}
\end{teaserfigure}

\newcommand{\refSuppImplementation}{Suppl. Sec. 1}
\newcommand{\refSuppHyperparams}{Suppl. Sec. 1.1}
\newcommand{\refSuppRendering}{Suppl. Sec. 2.1}
\newcommand{\refSuppTaskDescr}{Suppl. Sec. 2.2 }

\newcommand{\refSuppFigBaselines}{Suppl. Fig. 2}
\newcommand{\refSuppFigRosenbrock}{Suppl. Fig. 1}
\newcommand{\refSuppFigSamples}{Suppl. Fig. 3}
\newcommand{\refSuppFigODE}{Suppl. Fig. 4}

\begin{abstract}
Gradient-based optimization is now ubiquitous across graphics, but unfortunately can not be applied to problems with undefined or zero gradients.
To circumvent this issue, the loss function can be manually replaced by a ``surrogate'' that has similar minima but is differentiable.
Our proposed framework, \emph{\name}, automates this process by learning a neural approximation of the objective function, which in turn can be used to differentiate through arbitrary black-box graphics pipelines.
We train the surrogate on an actively smoothed version of the objective and encourage locality, focusing the surrogate's capacity on what matters at the current training episode.
The fitting is performed online, alongside the parameter optimization, and self-supervised, without pre-computed data or pre-trained models.
As sampling the objective is expensive (it requires a full rendering or simulator run), we devise an efficient sampling scheme that allows for tractable run-times and competitive performance at little overhead. 
We demonstrate optimizing diverse non-convex, non-differentiable black-box problems in graphics, such as visibility in rendering, discrete parameter spaces in procedural modelling or optimal control in physics-driven animation. 
In contrast to \changed{other derivative-free} algorithms, our approach scales well to higher dimensions, which we demonstrate on problems with up to 35k interlinked variables.
\end{abstract}

\maketitle

\acresetall

\newcommand\mycite[1]{\citeauthor{#1}, \citeyear{#1}}

\mysection{Introduction}{Introduction}
Gradient-based optimization has recently become an essential part of many graphics applications, ranging from rendering to find light or reflectance \cite{fischer2022metappearance, rainer2019neural, gardner2019deep}, over procedural material modelling \cite{hu2022node} to animation of characters or fluids \cite{habermann2021real, schenck2018spnets}.
These methods provide state-of-the-art results, in particular when combined with large amounts of training data and \acp{NN} that represent the desired mappings or assets.
In order to train these methods via \ac{GD}, the pipeline needs to be fully differentiable, allowing to backpropagate gradients from the objective function to the optimization parameters.
This often is enabled by intricate, task-specific derivations \cite{loubet2019reparameterizing, li2018differentiable} or requires fundamental changes to the pipeline, such as the switch to a dedicated programming language \changed{\cite{bangaru2021systematically, hu2019difftaichi, bangaru2023slangd, jakob2022dr}}.

In practice, the application of these ideas remains limited, as many existing graphics pipelines are black-boxes (\eg entire 3D modelling packages such as Blender or rendering pipelines such as Renderman \cite{christensen2018renderman} or Unity \cite{haas2014history}) that do not provide access to their internal workings and hence cannot be differentiated.
Further, even if we were given access to the pipeline's internals, the employed functions might not be differentiable (\eg the step function) or provide gradients that are insufficient for convergence \cite{metz2021gradients}.
The mindset of this article is that we only have access to a forward model, \eg a modeling pipeline, and a reference, \eg a target image. 
Using these two, a loss can be computed -- its gradients, however, cannot be used.

If a loss is not differentiable in practice, it can be approximated by a ``surrogate'' loss (\refFig{GradientFlow}).
The surrogate is a function that has similar minima as the true loss, but also provides gradients that are useful when employed in an optimization.
While the concept of surrogate modelling is not new (see \refSec{PreviousWork}), it remains unclear how to efficiently find a surrogate loss for any arbitrary graphics pipeline, as here the sampling must be sparse (recall, rendering a sample is expensive), 
\changed{and the optimization problems' dimensionality can vary by several orders of magnitude}.
In this article, we propose \emph{\name}, a systematic and efficient way of learning local surrogate losses, requiring no more than a forward model and a reference.
\myfigure{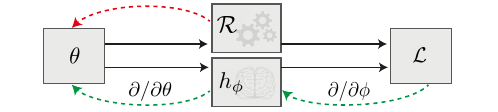}{Regular forward models $\mathcal{R}$ might not be able to provide gradients w.r.t. their input parameters \parameters (red arrow, top). Our approach, \name, provides this ability via a local learned surrogate \surrogate (green arrows, bottom) that maps \parameters to the associated loss and can be differentiated analytically.}

To learn our surrogate loss and use it in an optimization, we follow four key steps: 
We first smooth the original loss by convolving the cost landscape with a blur kernel, so that it 
provides gradients that lead to (improved) convergence when used during optimization.
We secondly fit a surrogate, a parametric function such as a neural network or a quadratic potential, to that smoothed loss.
As the surrogate is differentiable by construction, we can query it to get \emph{surrogate gradients} that drive the parameter optimization.  
We thirdly constrain this fit to the local neighbourhood of the current parameters, as the global cost landscape is in large parts irrelevant for the current optimization step.
By locally updating our surrogate, we allow it to focus on what matters ``around'' the current solution.
However, querying the objective function to create samples for our surrogate fitting is expensive, as each sample requires the execution of the full forward model, such as a light transport simulation or physics solver. 
Therefore, fourth and finally, we derive an efficient sampler that reduces the variance of the surrogate's gradient estimates
and thus allows us to use \name with a low number of sparse samples at tractable runtimes. 

In contrast to prior work \cite{hu2022inverse}, our local surrogate losses can be trained \emph{online}, alongside the actual parameter optimization, and \emph{self-supervised}, without the need for pre-trained models or pre-computed ground truth gradients. 
Moreover, in comparison to other gradient estimation techniques \cite{fischer2022plateau, spall1992multivariate}, our neural proxy allows us to move the noise and variance in the gradient estimates from the \emph{parameter} domain into the \emph{proxy} domain, where it is naturally smoothed by the network's hysteresis, allowing \name to scale up to very high dimensions.
In summary, our contributions are: 
\begin{itemize}
    \item \name, a framework that maps a forward model without usable gradients into a smooth, differentiable surrogate function, such as an \ac{NN}, with analytic gradients.
    \item Reducing the variance of surrogate and parameter updates to allow tractable runtimes and unsupervised and successful on-the-fly optimization. 
    \item Optimizing several non-differentiable black-box graphics problems from rendering (visibility) over modelling (discrete procedurals) to simulation and animation (optimal control).
    \item Showing that our surrogates scale favourably to higher dimensions, with up to several thousand correlated optimization variables, where existing derivative-free methods often struggle to converge.
    \item A publicly available implementation of our method and benchmarks at \changed{\url{https://github.com/mfischer-ucl/zerograds}}
\end{itemize}

We would like to emphasize that we do not claim superiority (less variance, better convergence, ...) over existing, specialized methods such as Mitsuba \cite{jakobmitsuba}, Redner \cite{li2018differentiable} or PhiFlow \cite{holl2020learning}, but rather broaden the toolkit  of inverse graphics solvers by now enabling gradient-based optimization on arbitrary, non-differentiable graphics pipelines.

\mysection{Previous Work}{PreviousWork}

\myparagraph{Optimization in Graphics}
Parameters of graphics models are now routinely optimized so as to fulfill user-provided goals.
The two main ingredients enabling this are gradient-based optimization and tunable architectures.
We will not consider the many different exciting architectures in this work, but focus on the optimization itself.
Gradients are typically computed by using a language that allows efficient auto-differentiation, such as PyTorch \cite{paszke2017automatic} or JAX \cite{jax2018github}, often targeting GPUs.
Unfortunately, several problems in graphics are not differentiable.

Gradient-free optimization algorithms \cite{rios2013derivative} such as asking a user \cite{marks1997design},
global optimization \cite{jones1998efficient}, direct search \cite{powell1994direct}, \ac{SA} \cite{kirkpatrick1983optimization}, \ac{SPSA} \cite{spall1992multivariate}, particle swarms \cite{parsopoulos2002recent} or \acp{GA} \cite{holland1973genetic} have largely fallen from favour in everyday graphics use.
This is partially due to the fact that gradient-free optimization -- even on smooth problems -- often requires a large number of function evaluations before converging \cite{jamieson2012query}, and, in general, struggles with convergence as problem dimensionality increases.
Additionally, gradient-free optimizers often suffer from high per-iteration cost in higher dimensions (e.g., \ac{FD} or \cite{zhang2020scalable}, see \refTab{comparison}),
require the computation of the Hessian matrix in Newton-type methods, or make use of covariance-matrices (CMA-ES \cite{hansen2016cma}), whose memory requirements grow quadratically with problem dimension and require computationally complex steps like inversion or eigendecomposition. 
The aforementioned aspects often lead to trade-offs between performance, scalability and runtime that practitioners have to take into consideration.
\ac{GD}, in contrast, has strong convergence guarantees (under convex objectives), is highly scalable, can be parallelized effectively, and has shown superior performance in high-dimensional settings (e.g., \ac{NN} training).
Unfortunately, it cannot be used on many relevant problems (although the cost landscape itself might be smooth), as many graphics pipelines are simply not designed to be differentiable.
Our approach circumvents this issue by fitting a model of the cost landscape, which can then be differentiated to provide \emph{surrogate} gradients that \ac{GD} can work with.
Even for non-smooth problems, our formulation makes the problem smooth and hence amenable to \ac{GD}. 

\begin{table}[t]
\centering
\caption{Comparison of optimization algorithms. $n$ is the problem dimensionality, $p$ is the population size in evolutionary algorithms, $k$ is the state's size in stateful algorithms, $b$ is the batchsize in multi-sample algorithms, $t$ and $t_i$ are the times required for a function evaluation and a state update, respectively. Stateful denotes whether the method maintains a state other than the current optimization parameter.}
\resizebox{\columnwidth}{!}{%
\begin{tabular}{@{}lcccccc@{}}
\toprule
     & Memory & Gradient & Iter. Cost & Scalable & Robust & Stateful \\ \midrule
Ours & \(n + k\)  & \cmark      & \(b \cdot t + t_\mathrm{GD}\) & \cmark & \cmark       & \cmark   \\
SPSA & \(n\)      & \cmark      & \(2t\)      & \cmark & \cmark       & \xmark   \\
FR22 & \(n\)      & \cmark      & \(b \cdot t\)      & \cmark & ?            & \xmark   \\
CMA-ES  & \(n^2\) + k & \xmark      & \(p \cdot t + t_\mathrm{CMA}\) & \cmark & \cmark & \cmark   \\
GA   & \(n\) & \xmark   & \(p \cdot t + t_\mathrm{GA}\)      & \xmark & \xmark       & \cmark   \\
FD   & \(n\)      & \cmark      & \(2n \cdot t\)     & \xmark & \xmark       & \xmark   \\
SA   & \(n\)      & \xmark      & \(t\)      & \xmark & \xmark       & \xmark   \\\bottomrule
\end{tabular}
}
\label{tab:comparison}
\end{table}

\newpage
\myparagraph{Differentiation}
Most graphics pipelines used in production (e.g., Blender, GIMP, Photoshop) are not differentiable, as they are not implemented in a differentiable programming language. The deeper underlying mathematical problem is that their output often relies on integration -- however, differentiation and the typical integral estimation through \ac{MC} cannot be interchanged without further considerations.
A prominent example are discontinuities in rendering, which have sparked a body of work by \citet{lee2018reparameterization,bangaru2021systematically,loper2014opendr,kato2018neural,li2018differentiable,liu2019soft, rhodin2015versatile, xing2022differentiable} or \citet{loubet2019reparameterizing}, to only name a few salient ones.
Similar problems appear in vector graphics \cite{li2020differentiable}, \acp{SDF} \cite{vicini2022differentiable, bangaru2022differentiable}, entire programs
\cite{chandra2021puzzles} or physics \cite{hu2019difftaichi,chang2016compositional,mrowca2018flexible}.
All these approaches require access to the internals of the graphics pipeline in order to replace or change parts such that gradients can be backpropagated.
Our approach, in contrast, assumes the pipeline to be a black-box and does not make any assumptions or approximations to the internals.

\changed{
\myparagraph{Gradient Smoothing} Another approach for optimizing non-diff\-erentiable problems has been proposed as \emph{stochastic} optimization.
Here, discontinuities and plateaus are smoothed out by optimizing over the expected value of a distribution (generally \ac{MC}-approxim. via randomized sampling) instead of a rigid parameter \cite{berthet2020learning, chaudhuri2010smooth, duchi2012randomized, staines2013optimization}, with the resulting gradient sometimes being referred to as \emph{zeroth-order} \cite{suh2022differentiable}.  
This has successfully been used to smooth out plateaus and discontinuities in rendering (\cite{fischer2022plateau, le2021differentiable}), contact dynamics in robotics (\cite{suh2022bundled, montaut2023differentiable}) and policy optimization in reinforcement learning (\cite{williams1992simple, suh2022differentiable}), albeit without the explicit learning of a proxy cost model.
}

\myparagraph{Proxies}
A key insight is that we only need the loss' gradients, and not those of the entire pipeline. 
Hence, if a part in a conventional graphics pipeline cannot be differentiated, we search for a similar function that we can differentiate instead, our \emph{proxy}.
As the proxy is an analytic function, the gradients w.r.t. its input can readily be computed via \ac{AD} and then be used for optimization.
Graphics is a good fit for neural proxies, as we can freely sample the objective in many cases, \eg by rendering an image or running a simulation (a ``simulation-optimization'' setting \cite{law2007simulation}).
While easy to do, creating a sample is expensive.

The concept of ``Neural Proxies'' was pioneered for physics by \citet{grzeszczuk1998neuroanimator} and is now applied to problems such as material editing
\cite{hu2022node}, photo editing \cite{tseng2022neural,fischer2020nicer}, 
hardware and design \cite{tseng2019hyperparameter}, software synthesis \cite{esmaeilzadeh2012neural}, simulation \cite{munk2019deep, renda2020difftune, ansari2022autoinverse} or  animation \cite{shirobokov2020black,navratil2019accelerating,sin2021neural}.
Rendering itself becomes differentiable when replaced by a \ac{NN} proxy
\cite{nalbach2017deep},
however, having a \ac{NN} emulate the complex behaviour of a full graphics pipeline might not scale to complex assets.

\myparagraph{Surrogate losses}
Surrogate losses \cite{queipo2005surrogate} (sometimes also called meta-models \cite{barton1994metamodeling, box1987empirical}) extend the idea of proxies by providing an approximation to the \emph{entire} forward model's behaviour by only emulating its response (or loss landscape), without necessarily replicating all the internals of the mdoel.
Differentiating the surrogate will provide gradients which can then be used for (gradient-based) optimization.
Surrogates are especially popular when taking a sample is expensive, like in airplane design \cite{forrester2009recent} or neural architecture search \cite{zhou2020econas}, and can be modelled in a number of ways, e.g., through polynomials \cite{jones1998efficient}, \acp{RBF} \cite{gutmann2001radial} and recently neural networks \cite{grabocka2019learning,patel2020learning}.
Most of these methods learn surrogates for the entire cost landscape (typically in a simplified setting, e.g., classification), with the exception of \acp{RSM}, which fit a first- or second order polynomial to the local neighbourhood, but are known to not converge on higher-dimensional problems \cite{wan2005simulation}.
More crucially, aside from global fitting, most methods assume the availability of a large set of data samples, e.g., a curated image collection like ImageNet \cite{deng2009imagenet}. 
In our setting, in contrast, sampling is expensive, which is why we sample \emph{sparsely and locally} and fit the surrogate in-the-loop, during optimization.
Our method hence extends the family of surrogate-based optimizers, and, in contrast to previous work, scales to a wide range of optimization tasks in high dimensions.

\mycfigure{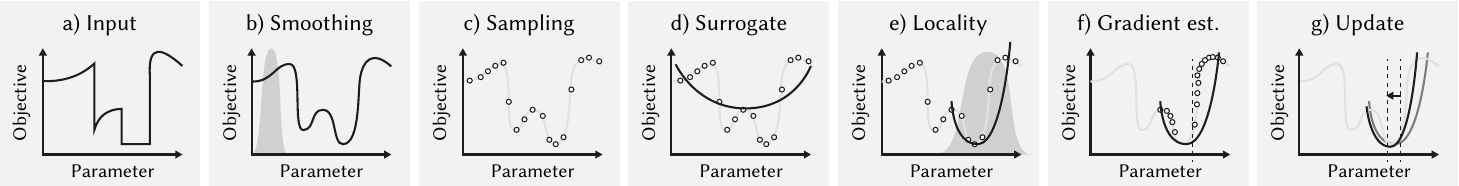}{A conceptual illustration of our approach; each subplot shows a one-dimensional cost landscape. For details please refer to \refSec{Method}: Overview.}

\myparagraph{Gradient estimation}
The surrogate itself is updated gradient-based, by sampling the objective function a finite number of times and then estimating the surrogate's gradient from its prediction error.
For general learning, building gradients is fundamentally a \ac{MC} estimation problem  \cite{mohamed2020monte}, akin to what graphics routinely is solving for rendering \cite{veach1998robust}.
We identify the similarity of estimating proxy gradients and simulating light transport (high dimensionality, sparsity, product integrands) and employ variance reduction based on importance sampling \cite{veach1998robust, kahn1950random, kajiya1986rendering} the local parameter neighbourhood to increase the efficiency of our surrogate gradient estimates. 

\mysection{Our approach}{Method}

\newcommand{\grad}[2]{\frac{\partial #1}{\partial #2}}

Given a scalar objective function $\objective(\parameters)\!: \parameterSpace\rightarrow\mathbb R_0^+$ over an \parameterDimension-dimen\-sional parameter space \parameterSpace, we would like to find the optimal parameters $\parameters^*$ that minimize \objective.
Typically, gradient-based optimizers such as SGD or ADAM are used for such a task.
However, in the setting of this work, their direct use is not possible, as the objective's gradients $\partial\objective/\partial\parameters$ are either not accessible (in a black-box pipeline), undefined (at discontinuities), zero (on a plateau) or too costly to compute (\eg when appearing in an integral). 
We are, however, able to sample this objective function by sampling a set of parameters and then comparing the resulting output with the reference. 
We propose to now \emph{locally} fit a tunable and differentiable surrogate function $\surrogate(\parameters, \surrogateParameters)$ to those samples, whose derivative $\partial\surrogate/\partial\parameters$ will act as \emph{surrogate gradient} and drive the optimization. 

\paragraph{Overview}

Our approach is summarized in \refAlg{Method} and \refFig{Concept}.
Given an arbitrary (potentially non-smooth and non-convex) objective function (\refFig{Concept}a, also called loss landscape) and a randomly initialized parameter state \parameters, we first smooth the objective via convolution with a Gaussian kernel in order to reduce plateaus (\refSec{Smoothing} and \refFig{Concept}b).
We subsequently fit our surrogate \surrogate (\refSec{Surrogate}) to this smooth objective. 
However, sampling is expensive (requiring a full rendering or simulator run), and global sampling and surrogate fitting would be very approximate (\refFig{Concept}c, d), which is why we enforce locality via another Gaussian kernel (\refFig{Concept}e and \refSec{Locality}) and hence encourage the surrogate to focus on what matters at the current optimization iteration. 
Unfortunately, we do not have supervision gradients to train our surrogate on, which is why we must estimate the surrogate's gradient (\refSec{Estimator}).
While some samples are more informative than others, it is unclear how to find those, i.e., how we can efficiently sample this convolved space. 
To this end, we derive an efficient importance-sampler (\refFig{Concept}f and \refSec{Sampling}) that samples according to the locality terms and thus reduces the variance of the estimated surrogate gradient. 
Finally, we use this estimated gradient to update our surrogate (\refFig{Concept}g) in order to improve its fit to the objective. 
We can then descend along the surrogate's gradient $\partial \surrogate / \partial \parameters$ (readily available via \ac{AD}) to update the optimization parameter \parameters and repeat the process from b) onward, i.e., the surrogate is updated from its previous state instead of re-fit. 
Lines \ref{lst:opt1} and \ref{lst:opt2} in \refAlg{Method} illustrate this, where \textsc{Optimize} performs gradient descent steps on a variable. 

\begin{algorithm}[b]
    \caption{High-level pseudo-code of \name.}
    \begin{algorithmic}[1]
    \Statex \textbf{Input:} objective $\objective$, surrogate $\surrogate$
    \Statex \textbf{Output:} optimized parameters \parameters minimizing \objective
    \Procedure{\name}{%
        \objective, %
        \surrogate}
    \State $\surrogateParameters$ := \Call{Init}{\null} \Comment{surrogate parameters}
    \State $\parameters$ := \Call{Uniform}{\null} \Comment{optimization parameters}
    \For{i}
        \State \surrogateParameters = 
        \Call{Optimize$_1$}{%
        \surrogateParameters, 
        \Call{EstimateGradient}{\surrogateParameters, \parameters \objective, \surrogate}} \label{lst:opt1}
        \State \parameters = \Call{Optimize$_2$}{%
        \parameters, 
        $\partial \surrogate / \partial \parameters$}
        \label{lst:opt2}
    \EndFor
    \State \Return $\parameters$
    \EndProcedure
    \end{algorithmic}
    \label{alg:Method}
\end{algorithm}

\mysubsection{Smooth objective}{Smoothing}
As the objective might not always be differentiable (or provide gradients that are of little use \cite{metz2021gradients}), we seek to find a function that has similar optima and is differentiable.
In practice, the issue is not so much in non-differentiable point singularities (which are even present in the popular ReLU activation), but regions with zero gradients (``plateaus'').
These can be removed by convolving the objective with a blur kernel.
Similar ideas have been applied to rasterization \cite{liu2019soft, petersen2022gendr} and path-tracing \cite{fischer2022plateau}, which we scale to arbitrary spaces.
We define the smooth objective 
$
\smoothObjective(\parameters)\!:
\parameterSpace\rightarrow\mathbb R_0^+
$
 as
\begin{align}
\label{eq:SmoothObjective}
\smoothObjective(\parameters) = 
\smoothingKernel * \objective(\parameters)=
\int_\Theta
\smoothingKernel(\offset)
\objective(\parameters-\offset)
\mathrm d\offset
,
\end{align}
a convolution of the objective \objective and a smoothing kernel \smoothingKernel, which we choose to be a Gaussian. 
Convolution with a Gaussian kernel has several desirable properties, e.g.,
convexity is preserved, it holds that 
$L_g \leq L_f$, i.e., the smooth objective is stronger Lipschitz-bound than \objective, and the gradient $\nabla_g$ is Lipschitz-continuous even when $\nabla_f$ is not \cite{nesterov2017random}, as is the case on many of our problems (e.g.,  \refFig{Concept}a)). We show visualizations of \refEq{SmoothObjective} in \refFig{blur_visualization}.

\mysubsection{Surrogate}{Surrogate}
The key ingredient, the surrogate $
\surrogate(\parameters, \surrogateParameters)\!:
\parameterSpace \,
\times \,
\surrogateParameterSpace
\rightarrow\mathbb R_0^+
$
, consumes \parameters (like \objective and \smoothObjective, which it emulates), but also takes the tunable parameters \surrogateParameters from the \surrogateParameterDimension-dimensional surrogate parameter space
\surrogateParameterSpace.

We encode our surrogate in a differentiable proxy function of variable form, which can take the form of polynomials, \acp{RBF} or \acp{NN} (see \refSec{Results} for examples) and whose analytic parametrization allows to easily get the surrogate gradient $\partial\surrogate/\partial\surrogateParameters$ and the parameter gradient 
$\partial\surrogate/\partial\parameters$ via automatic differentiation.
In contrast to linear methods (e.g., \cite{zhang2020scalable, fischer2022plateau, spall1992multivariate}), our continuous surrogate formulation allows us to perform one or more gradient descent steps on the surrogate and to evaluate the estimated loss surface at a new position without re-running the forward model.

\mysubsection{Localized surrogate loss}{Locality}

Matching \surrogate to \smoothObjective across the entire domain \parameterSpace might be too ambitious and furthermore is unnecessary, as most first-order gradient-based optimizers only ever query values at or around the current parameter \parameters.
Instead, we create a surrogate that is focused around the current parameters 
by locally sampling the objective function.

The loss of the surrogate parameters \surrogateParameters ``around'' \parameters hence is
\begin{align}
\label{eq:LocalizedLoss}
\surrogateLoss(\parameters, \surrogateParameters)
=
\int_\parameterSpace
\locality(\lossOffset, \parameters)
\left(
\smoothObjective(\lossOffset) 
- 
\surrogate(\lossOffset, \surrogateParameters)
\right)^2
\mathrm d\lossOffset,
\end{align}
where \locality is a weighting function that chooses how much context is considered around the current solution and \lossOffset is from the parameter space \parameterSpace.
We again choose a Gaussian with mean \parameters here, which is not to be confused with the smoothing kernel \smoothingKernel.
\refEq{LocalizedLoss} illustrates how the surrogate never has access to the gradients of the true objective -- these might not even exist --, but learns self-supervised by only sampling the (smoothed) loss \smoothObjective.

\mysubsection{Estimator}{Estimator}
Combining the smoothed loss in \refEq{SmoothObjective} and the localized surrogate loss in \refEq{LocalizedLoss}, we arrive at the following expression: 
\mymath{\lossIntegrand}{I}
\begin{equation}
\label{eq:surrogateloss}
\surrogateLoss(\parameters, \surrogateParameters)
=
\int_\parameterSpace
\underbrace{
\locality(\lossOffset, \parameters)
\left(
\left[
\int_\Theta
\smoothingKernel(\offset)
\objective(\lossOffset-\offset)
\mathrm d\offset
\right]
- 
\surrogate(\lossOffset, \surrogateParameters)
\right)^2}_{:= \lossIntegrand(\lossOffset, \, \surrogateParameters)}
\mathrm d\lossOffset.
\end{equation}

We are now interested in the gradient of this expression w.r.t. the surrogate parameters \surrogateParameters, \ie 

\begin{equation}
\label{eq:leibnizGradients}
\frac{\partial}{\partial \surrogateParameters}
\surrogateLoss(\parameters, \surrogateParameters)
 =
\frac{\partial}{\partial \surrogateParameters} 
\int_\parameterSpace 
\lossIntegrand(\lossOffset, \surrogateParameters) 
\mathrm d\lossOffset  
=  
\int_\parameterSpace
\frac{\partial}{\partial \surrogateParameters}
\lossIntegrand(\lossOffset, \surrogateParameters) 
\mathrm d\lossOffset
.
\end{equation}

The above equality holds, according to Leibniz' rule of differentiation under the integral sign, if, and only if, the integrand is continuous in \surrogateParameters and \lossOffset \cite{li2018differentiable}. 
Our Gaussian locality weight \locality fulfills this and  $\surrogate(\lossOffset, \surrogateParameters)$ is continuous by definition, as it is a \ac{NN} or quadratic potential. 
Through our previously introduced convolution (\refEq{SmoothObjective}), the -- originally discontinuous -- objective \objective becomes smooth and hence leads to the inner integral being continuous in \lossOffset.  
Leveraging the fact that a composition of continuous functions also is continuous,
we can say that $I(\lossOffset, \surrogateParameters)$ is continuous in \surrogateParameters and hence use \refEq{leibnizGradients} as our gradient estimator:
\begin{align}
\frac{\partial}{\partial \surrogateParameters}\surrogateLoss &=
\int_\parameterSpace 
\frac{\partial}{\partial \surrogateParameters} \locality(\lossOffset, \parameters)
\left(
\left[
\int_\Theta
\smoothingKernel(\offset)
\objective(\lossOffset-\offset)
\mathrm d\offset
\right]
- 
\surrogate(\lossOffset, \surrogateParameters)
\right)^2
\mathrm d\lossOffset
\\ 
& 
\label{eq:gradientIntegral}= 
\int_\parameterSpace 
2 \locality(\lossOffset, \parameters) 
\frac{\partial \surrogate(\lossOffset, \surrogateParameters)}{\partial \surrogateParameters} 
\left( 
\surrogate(\lossOffset, \surrogateParameters) 
- 
\int_\Theta
\smoothingKernel(\offset)
\objective(\lossOffset-\offset)
\mathrm d\offset
\right)
\mathrm d\lossOffset.
\end{align}

\mymath{\integralResult}{\mathcal I}
\mymath{\innerSampelCount}N
\mymath{\outerSampelCount}M

Here, the inner integral over \offset is conditioned on the outer variable \lossOffset, leading to a nested integration problem. 
\changed{
A general, unbiased solution would estimate the inner and outer integrals with \innerSampelCount and \outerSampelCount samples, respectively, where $\outerSampelCount \propto \innerSampelCount$, leading to quadratic complexity.
However, in this special case the function acting on the inner integral is linear, and the nested estimator thus remains unbiased even for constant \innerSampelCount (see \cite{rainforth2018nesting}, Sec. 5 \& Fig. 2). 
}

We now aim to re-arrange \refEq{gradientIntegral} into a double-integral of a product, a form that is reliably solvable by the well-known approaches that solve the rendering equation \cite{veach1998robust}.
We hence write \refEq{gradientIntegral} as 
\begin{align}
\begin{aligned}
\label{eq:doublegradientIntegralLong}
\frac{\partial}{\partial \surrogateParameters}\surrogateLoss 
= 
\int_\parameterSpace 
&\int_\Theta
2\locality(\lossOffset, \parameters) 
\frac{\partial \surrogate(\lossOffset, \surrogateParameters)}{\partial \surrogateParameters} 
\surrogate(\lossOffset, \surrogateParameters)\mathrm d\offset
-  \\ 
&\int_\Theta
2\locality(\lossOffset, \parameters) 
\frac{\partial \surrogate(\lossOffset, \surrogateParameters)}{\partial \surrogateParameters} 
\smoothingKernel(\offset)
\objective(\lossOffset-\offset)
\mathrm d\offset \, 
\mathrm d\lossOffset, 
\end{aligned}
\end{align}
where we use the fact that integrating an expression that is independent of the integration variable (here \offset) reduces to multiplication by the volume of the integration domain, which we here assume to be normalized to unit volume, i.e., $\int_\parameterSpace d\offset=1$. 
Now the two inner integrals can be written as one, and factored as
\begin{equation}
\label{eq:doublegradientIntegral}
\frac{\partial}{\partial \surrogateParameters}\surrogateLoss 
= 
\iint_\parameterSpace 
2\locality(\lossOffset, \parameters) 
\frac{\partial \surrogate(\lossOffset, \surrogateParameters)}{\partial \surrogateParameters} 
\left( 
 \surrogate(\lossOffset, \surrogateParameters)
- 
\smoothingKernel(\offset)
\objective(\lossOffset-\offset)
\right) 
\mathrm d\offset
\mathrm d\lossOffset
.
\end{equation}

This integral spans the product space $\parameterSpace \times \parameterSpace$ and has the following unbiased estimator with linear complexity $\mathcal{O}(\innerSampelCount)$:
\begin{align}
    \label{eq:estimator}
    \frac{\partial}{\partial \surrogateParameters}\surrogateLoss
    &\approx 
    \frac{1}{\innerSampelCount}
    \sum_i^\innerSampelCount 
    \frac{2\locality(\lossOffset_i, \parameters)}{p(\lossOffset_i, \offset_i)}
    \frac{\partial \surrogate(\lossOffset, \surrogateParameters)}{\partial \surrogateParameters}
    \left(
    \surrogate(\lossOffset_i, \surrogateParameters)
    -
    \smoothingKernel(\offset_i)\objective(\lossOffset_i-\offset_i) \right)
    ,
    \\ 
    &= 
    \label{eq:estimator_ad}
    \frac{\partial}{\partial \surrogateParameters}
    \frac{1}{N}
    \sum_i^\innerSampelCount 
    \frac{\locality(\lossOffset_i, \parameters)}{p(\lossOffset_i)p(\offset_i)} \left(
    \surrogate(\lossOffset_i, \surrogateParameters)
    -
    \smoothingKernel(\offset_i)\objective(\lossOffset_i-\offset_i)
    \right)
    ^2
    .
\end{align}

It seems somewhat contrived to go through all the above reformulations to arrive from \refEq{leibnizGradients} at \refEq{estimator}. 
Note, however, that differentiating \emph{inside} the integral, enabled by \refEq{leibnizGradients}, removes the non-linear function acting on the inner integral in \refEq{surrogateloss} and hence enables us to formulate an \emph{unbiased} estimator of the smoothed, localized loss in $\mathcal{O}(\innerSampelCount)$.
This re-formulation is only possible in special cases (here, for a second-order polynomial acting on the inner integral, see \cite{rainforth2018nesting} Sec. 4) and would not be possible for other popular distance measures between \surrogate and \objective, e.g., the \ac{KL}-divergence or the Hinge- or Exponential-losses, which would introduce bias into the optimization due to non-linearities in their derivatives \cite{deng2022reconstructing, nicolet2023recursive}.

\mymath{\surrogategradient}{\nabla \surrogateParameters}
\begin{algorithm}[tb]
    \caption{Estimating the surrogate's gradient}
    \label{alg:EstimatorGradient}
    \begin{algorithmic}[1]
    \Statex \textbf{Input: } surrogate parameters \surrogateParameters, optimization parameter \parameters,
    \Statex \hspace{\algorithmicindent}\hspace{\algorithmicindent} objective \objective, surrogate \surrogate
    \Statex \textbf{Output:} surrogate gradient $\surrogategradient$
    \Procedure{EstimateGradient}{\surrogateParameters, \parameters \objective, \surrogate}

        \State \surrogategradient\! := 0
        
        \For{\sampleCount}

        \State $\lossOffset, \offset$ :=
        \Call{Normal}{\spreadouter}, \Call{Normal}{\spreadinner}
      
        \State $s$ := $
            (\surrogate(\lossOffset, \surrogateParameters) - 
            \objective(\lossOffset-\offset))^2$

        \State \surrogategradient += \Call{Grad}{s, \surrogateParameters}

        \EndFor
        \State \Return $\surrogategradient / \sampleCount$
    \EndProcedure   
    \end{algorithmic}
\end{algorithm}

\mycfigure{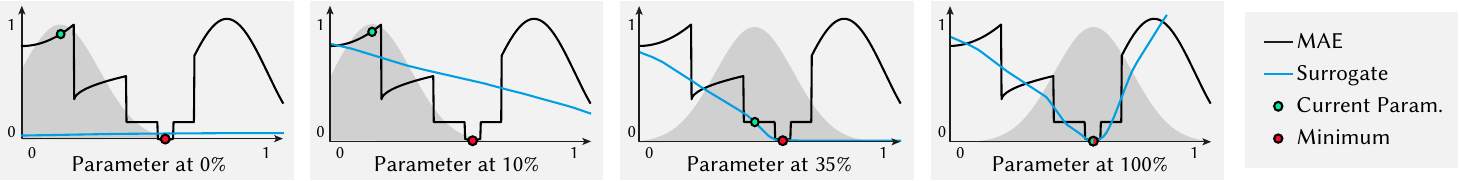}{
\changed{A 1D-example with the function from \refFig{Concept}a) and our neural surrogate (blue), which learns a local approximation of the loss (black, MAE) and provides gradients for the optimization parameter (green). The sampling distribution is displayed in grey,
state is shown at 0\%, 10\%, 35\% and 100\% total iterations.}
}
\mysubsection{Sampling}{Sampling}
For uniform random sampling, the \ac{MC} estimator in \refEq{estimator} will exhibit substantial variance, leading to slower convergence and hence longer runtimes. 
Instead, we would like to sample in a way that maximizes a sample's importance and therefore produces more meaningful gradients for the same sampling budget. 
Thanks to our reformulation of the nested integral, it is evident that the integration domain now is $\parameterSpace\times\parameterSpace$, and that the magnitude of the gradient -- the quantity we would like to estimate -- is determined by three factors: the difference between the surrogate's prediction and (smooth) objective \objective, and the locality terms $\locality$ and $\smoothingKernel$.

While we cannot trivially compute the \ac{PDF} of the surrogate's prediction error, 
our local surrogate formulation shows that we can reduce the variance by importance sampling \cite{veach1998robust} for both locality terms. 
This again allows us to focus our surrogate  on the regions of the parameter space that matter at the current optimization iteration while ignoring large amounts of space, which is especially helpful in higher dimensions. 
The parameter \spreadouter of the locality \locality determines how far the current solution's sampling radius is spread out, while \spreadinner determines the amount of smoothing, and is generally set to 15\% of \spreadouter (for details, please see \refSuppHyperparams). 
We display the resulting gradient estimator in \refAlg{EstimatorGradient}. 

\mysubsection{Summary}{Summary}
In combination, the above elaborations allow us to optimize the objective function $\objective(\parameters)$ through our surrogate's \emph{surrogate gradients} $\partial\surrogate/\partial\parameters$.
We emphasize that, in contrast to prior work \cite{hu2022node, patel2020learning, grabocka2019learning}, our surrogate is learned self-supervised, without any ground truth supervision in the form of pre-computed gradients, and is optimized on-the-fly, alongside the parameter \parameters. 
This is made possible by a low number of \changed{samples} which we achieve through our efficient estimators. 
As such, it allows the application of our method to systems where only a forward model is given. 
In the following sections, we will detail and evaluate some exemplary applications. 
\changed{
Further, we provide a simple, illustrative 1D-example and visualize our learned surrogate loss over the course of the optimization in \refFig{illustrative}.
}

\mysection{Evaluation}{Evaluation}
Our evaluation compares different methods on a range of tasks (see \refSec{Tasks} and \refSuppTaskDescr for detailed task descriptions): we validate our design choices through ablations on lower-dimensional tasks in \refSec{Tasks} and compare against established derivative-free optimizers on higher-dimensional, real-world tasks in \refSec{higherdimensions}.  
The reason for this is twofold: first, in the higher-dimensional regime, the task complexity makes it non-evident to see which of the ablated attributes lead to the method failing, and second, the derivative-free algorithms could solve some of the lower-dimensional tasks by simply brute-forcing the solution (which is a valid way of solving the problem, but besides the scope of interest here). 

As our objective \objective, we use the \ac{MSE} between the current rendered state and the target, if not otherwise specified.
For details on our proxy's architecture and hyperparameters, please see \refSuppImplementation. 

\mysubsection{Methods}{methods}
We compare our approach to several ablations and variants, out of which some correspond to existing published methods. 
All methods operate in \emph{image space} only and do not have access to any ground truth supervision or parameters.
We structure the space of methods by the type of i) smoothing, 
ii) surrogate, and 
iii) sampling.
Our full method, \method{Ours}, implements \refAlg{EstimatorGradient}: it smooths the loss via convolution with a Gaussian (\refEq{SmoothObjective}), uses a neural proxy and draws samples by importance-sampling the locality terms.
We compare our full approach to the following methods and ablations: 

\method{NoSmooth}, where we ablate the smoothing convolution (\refEq{SmoothObjective}) and directly sample the non-smooth loss.

\method{NoNN}, where we replace the \ac{NN} in the surrogate by a quadratic potential function of the form 
$(\inputvector, 1)^\intercal \potentialMatrix \, (\inputvector, 1)$, where, for an $n$-dimensional problem, $\potentialMatrix$ is a symmetric matrix in $\mathbb{R}^{n+1\times n+1}$. 

\method{NoLocal}, where we ablate the locality by drawing uniform random samples from the domain.

\method{FD}, our implementation of \acl{FD}, with an optimally chosen $\pm \epsilon$ in each dimension. 

\method{FR22}, which re-implements the approach presented by \citet{fischer2022plateau}, who derive gradients by stochastically perturbing the optimization parameters \parameters at every iteration and weighting the resulting loss values by a gradient-of-Gaussian kernel, effectively creating a linear gradient estimate akin to a stochastic multi-sample version of \citet{spall1992multivariate}. 

\mysubsection{Protocol}{Protocol}
\mymath{\ntasks}{10}
The results we report are the median values over an ensemble of \ntasks independent runs of random instances of each task.
To ensure fairness, all methods are run in their best configuration.
For each run, the parameter initialization and ground truth (where possible) are randomly re-sampled and the surrogate, optimizer and all other stateful components are re-initialized from scratch.

\newcommand{\tmpVal}[0]{00.0$\,\times$}
\newcommand{\boring}[1]{\textcolor{lightgray}{#1}}

\begin{table}[]
    \setlength{\tabcolsep}{3pt}
    \centering
    \caption{An overview of our method and its ablations and competitors. To the right, we show the competitors' relative error ratio at the iteration where our method achieves 95\,\% error reduction - i.e, how much others are behind.}
    \begin{tabular}{llllrrr}
        \multicolumn1l{Method}&
        \multicolumn1l{Smooth.}&
        \multicolumn1l{Surr. \surrogate}&
        \multicolumn1l{Sampler}&
        \multicolumn1l{Rend.}&
        \multicolumn1l{Model.}&
        \multicolumn1l{Anim.}\\
        \toprule
        \method{NoSmooth}&
        None&
        \boring{NN}&
        \boring{Gauss} &  1.2$\,\times$&
         3.9$\,\times$&
         1.5$\,\times$\\
        
        \method{NoNN}&
        \boring{Gauss}&
        Quad.&
        \boring{Gauss} & 8.9$\,\times$&
         792.4$\,\times$&
         16.3$\,\times$\\

        \method{NoLocal}&
        \boring{Gauss}&
        \boring{NN}&
        Uniform & 12.3$\,\times$&
         613.4$\,\times$&
         22.4$\,\times$\\

        \midrule
        \method{FD}&
        None&
        Linear&
        Box   & 24.5$\,\times$&
         654.3$\,\times$&
         10.2$\,\times$\\
        
        \method{FR22}&
        \boring{Gauss}&
        Linear&
        Import. & 11.0$\,\times$&
         323.6$\,\times$&
         3.3$\,\times$\\
        
        \midrule
        
        \method{Ours}&
        Gauss&
        NN&
        Gauss & \boring{1.0$\,\times$}  & \boring{1.0$\,\times$} & \boring{1.0$\,\times$}
        \\
        \bottomrule
    \end{tabular}
    \label{tab:overview}
\end{table}

\mysubsection{Tasks}{Tasks}
We validate our method on two sets of tasks: the first set consists of lower-dimensional tasks from rendering, animation, modelling and physics, all of which are not trivially amenable to gradient-based optimization due to (partially) discrete parameter spaces, discontinuous integrals or non-differentiable frameworks, and serves the purpose of evaluating our approach's design decisions. 
We provide a short description of each task in the following section and refer the reader to \refSuppTaskDescr \, for more details on task setup and to \refFig{results_qualitative} for task-specific visualizations.

Additionally, we evaluate how well our method scales to more realistic, higher-dimensional inverse optimization problems on a second set of tasks in \refSec{higherdimensions}.

\mysubsubsection{Differentiable Rendering}{DifferentiableRendering}
Discontinuities in rendering arise from the visibility function that appears inside the integral, in which case the gradient of the integral cannot be computed as the integral of the gradient.
Typical solutions include re-parametrization or edge-sampling for path tracing \cite{loubet2019reparameterizing, li2018differentiable} or replacing step functions with soft counterparts in rasterization \cite{petersen2022gendr, liu2019soft} and use framework-specific implementations.

\task{Cornell-Box} We optimize the light's horizontal and vertical translation and the axial rotation of both boxes inside the Cornell box.
The discontinuities arise from visibility changes at the silhouette edges of the moving boxes and light.

\task{BRDF} Here, we optimize the material properties (RGB reflectance and \ac{IOR}) of a material test-ball illuminated under environment illumination.  
Optimizing properties of ideal specular objects, such as their \ac{IOR}, often is challenging even for modern differentiable path tracers \cite{jakobmitsuba}.
Our method does not make any assumptions on the underlying function, so we can successfully optimize these cases, too. 

\task{Mosaic} In this task, we simultaneously optimize the vertical rotation of 320 cubes to match a reference.
Again, the discontinuities arise for pixels at the silhouette edges that change their color discontinuously with the occluding cube's rotation.

\mysubsubsection{Procedural Modelling}{ProceduralModelling}
Procedural modelling mainly uses no\-des from two categories, \emph{filtering} nodes (that are differentiable by construction) and \emph{generator} nodes, that operate on discrete parameters, such as a brick texture generator.
Node relations often are highly non-linear due to complex material graphs and their interplay with other pipeline parameters.
Moreover, the connections inside the nodegraph are a combinatorical problem with a highly discontinuous loss landscape. 
While previous research has made great progress in this field \cite{hu2022node, guerrero2022matformer, shi2020match, hu2022inverse, li2023end}, it still often either involves lengthy, framework-specific pre-training, or relies on the existence of differentiable material libraries.

\task{Wicker} A procedural modelling scenario where we simultaneously optimize the parameters of a node graph that creates a woven wicker material. We optimize the number of horizontal and vertical stakes and the number of repetitions across the unit plane.

\task{LED} Given a target setting, we optimize each LED panel in a digital display to either be on or off.
The display consists of 12 panels, each with 28 elements, leading to a 336-dim. binary problem. 

\task{Node-Graph} In this task, we directly optimize the connectivity of a material graph - a mixture of a combinatorical and procedural modelling problem.
We encode all possible edges for a given set of nodes in a connection matrix and optimize the matrix entries.
If an entry rounds to 1, the corresponding edge is inserted into the node graph, else removed.
Our test graph has 8 nodes (several inputs and outputs each, 24 valid connections and matrix entries).
Some graph edges constrain each other, e.g., when a shader node already is connected, a new connection will not result in an updated image, making this an even more inter-linked problem.

\mysubsubsection{Animation}{Animation}
Accurate animation often relies on differential equations to solve the underlying physics equations which govern a character's behaviour or movement.
Oftentimes, the forward model of such an animation is inherently discontinuous (for instance, due to collisions) while at the same time, the underlying physics solver is not differentiable, as its output is the solution of an integral approximated by discrete sampling locations (the time steps).
This is similar to the problem encountered in differentiable rendering: if the integrand (e.g., the time at which a force starts acting) is discontinuous, it is incorrect to simply differentiate the integral estimate to get a gradient estimate (see \refSuppFigODE). 

\task{Rocket} A physical simulation where we optimize the discrete event time at which a rocket's engine must be turned off in order to reach a certain target point.  
We simultaneously optimize 10 rockets flying in parallel.
As the forward model is solved with a finite number of time steps, a small change does not necessarily translate to a different outcome, leading to zero gradients almost everywhere.

\task{Gravity} This task runs a physics solver to infer the collision behaviour of three cubes that are dropped onto each other.
If optimized correctly, the cubes will form a tower after being dropped.
We optimize the two upper cubes' initial translation and their coefficients of restitution, the ``bounciness'', which we provide as the input to the solver. 

\mysubsection{Results}{Results}
We display the results of all methods listed in \refTab{overview} in \refFig{results}, with one subfigure per task, and show a quantitative analysis in the right part of \refTab{overview}.
We group the results into rendering (\task{Cornell Box}, \task{Brdf}, \task{Mosaic}), modelling (\task{Wicker}, \task{Node-Graph}, \task{Led}) and animation (\task{Rocket}, \task{Gravity}), which correspond to the three rightmost columns in \refTab{overview} and the rows in \refFig{results}, respectively. 

From the convergence plots in \refFig{results}, it becomes evident that \method{NoLocal} works in select cases, but often struggles to find the correct solution, especially where the domain is higher-dimensional (e.g., \task{Mosaic}), as a uniform random sampling of the parameter space introduces substantial variance in the gradient estimates. 
Similarly, \method{NoNN} is challenged in higher-dimensional cases and does not converge reliably, which we attribute to the reduced expressiveness of the quadratic potential. 
However, this shows that for simple tasks (e.g., \task{Cornell-Box}), the proxy does not need to be overly complicated. 
Finite Differences (\method{FD}) works well and makes steady progress towards the target, but does not scale well to higher dimensions, as an $n$-dimensional problem requires $2n$ function evaluations for a single gradient step (see the wall-time plots in \refFig{results} and \refTab{comparison}). 
\method{FR22} works reliably on all tasks, but often converges slower than our method.
Surprisingly, we find that ablating the inner smoothing operation in \method{NoSmooth} only slightly impedes performance (ca. 2x), which we partly attribute to the implicit smoothing introduced by the surrogate fit.
\myfigure{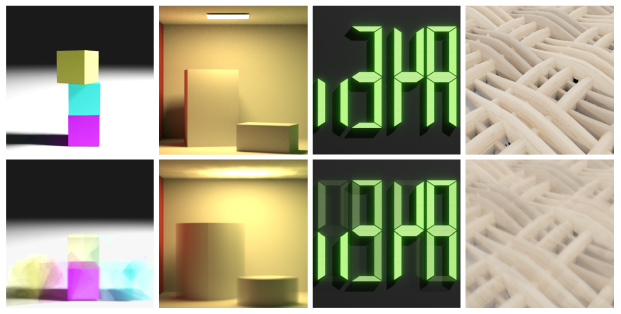}{Samples of the smooth objective (bottom row) on which we learn our surrogate: Perturbing the rigid scene parameters (top row) smooths discontinuities, e.g., the binary on/off for the \task{LED} task (an inset is shown).}
In almost all cases, \method{Ours} works best and faithfully recovers the true parameters. 
The overhead of our method compared to its competitors is small: for smoothing, it suffices to slightly perturb the current state, i.e.,
no additional evaluation of \objective is required. 
The \ac{NN} query is very efficient as it can be parallelized on the GPU, and the surrogates are relatively shallow (for implementation details see \refSuppImplementation), 
providing \method{Ours} with the best quality-speed relation, as is evident from the right bottom subfigure in \refFig{results}, where we show the (normalized and re-sampled) mean performance of all methods.

\mycfigure{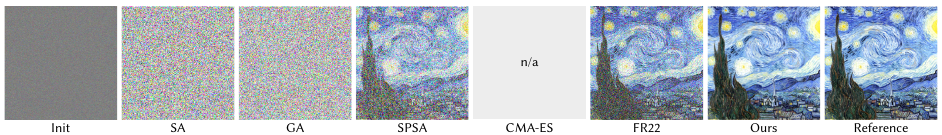}{We show an equal-sample comparison (i.e., the same budget of function evaluations) for the task of optimizing a $\texturedim \times \texturedim \times 3$ texture. CMA-ES cannot be run on this example due to its quadratic memory complexity causing out-of-memory errors on our 64 GB RAM machine.}

\mycfigure{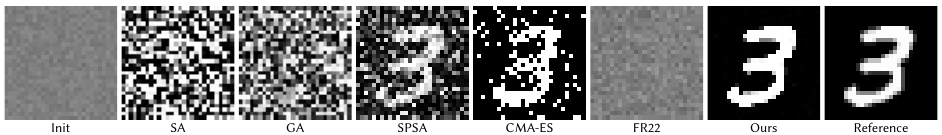}{We show an equal-sample comparison for the task of optimizing the \numweights weights of a \acs{MLP} such that it encodes digits from MNIST \cite{lecun1998mnist}.}

\mycfigure{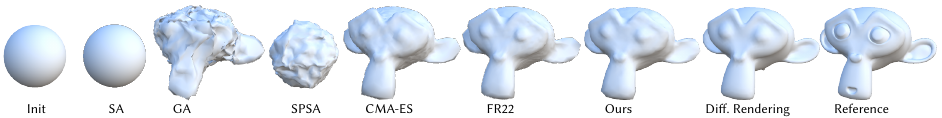}{We show an equal-sample comparison for the task of optimizing the 3D positions of a tessellated sphere with \numvertices vertices to match a rendered reference shape. ``Diff. Rendering'' uses the analytical gradients from \citet{Nicolet2021Large}.}

\mycfigure{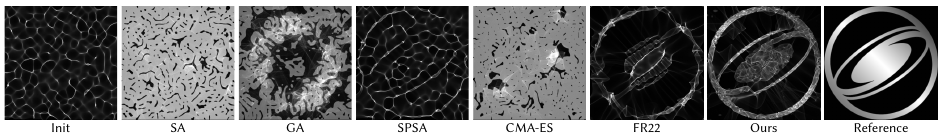}{We show an equal-sample comparison for the task of optimizing a 1,024-dim. heightfield such that the resulting caustic resembles the reference image.}

\mysubsection{Higher Dimensions}{higherdimensions}
We here evaluate our method on higher-dimensional problems from the inverse rendering literature and compare our approach against the established derivative-free optimizers genetic algorithms (\almostMethod{\ac{GA}}) \cite{holland1973genetic}, simulated annealing (\almostMethod{\ac{SA}}) \cite{kirkpatrick1983optimization}, simultaneous perturbation stochastic approximation (\almostMethod{\ac{SPSA}}) \cite{spall1992multivariate} and \almostMethod{CMA-ES} \cite{hansen2016cma}.
Due to the high dimensionality of these experiments, we found it necessary to increase the surrogate capacity and the gradient batchsize (for implementation details, please see \refSuppImplementation). 
All results we show are equal-sample comparisons, i.e., achieved after the same number of function evaluations, disregarding the fact that \almostMethod{CMA} requires significantly (multiple times) more runtime than all other approaches.
To avoid clutter in the main manuscript, we show the outcome of our ablated methods on theses tasks in \refSuppFigBaselines.

\task{Texture} First, we optimize the $\texturedim \times \texturedim$ RGB pixels of a texture in \refFig{texture}, a relatively simple task with a smooth cost landscape and no correlation between the optimization variables. 
Our surrogate gradients lead to a successful optimization outcome, while the derivative-free optimizers \almostMethod{\ac{GA}} and \almostMethod{\ac{SA}} fail to converge due to the high problem dimensionality. 
\almostMethod{CMA} cannot be run on our 64GB RAM machine because of the quadratic memory requirements of the covariance matrix.
Both \almostMethod{\ac{SPSA}} and \almostMethod{FR22} make progress towards the target, but require more time to converge. 

\task{MLP}
To increase the correlation between the optimization variables from the previous task, we repeat a similar experiment in \refFig{mlp_mnist}, where we use our method to optimize the weights of a \ac{MLP} such that it overfits single digits from the MNIST \cite{lecun1998mnist} dataset, i.e., learns a mapping from continuous 2D coordinates in (0,1) to a monochrome color value at the corresponding pixel location.
The results are similar: our method has already converged, while \almostMethod{\ac{SPSA}} and \almostMethod{CMA} make progress but require more function evaluations, and both \almostMethod{\ac{GA}} and \almostMethod{\ac{SA}} do not converge at all. 
Interestingly, \almostMethod{FR22} does not converge either.
This is in line with recent findings that show that perturbation-based methods do not perform well on emulating backpropagation in neural networks \cite{nesterov2017random, chandra2021unexpected, belouze2022optimization}.

\mycfigure{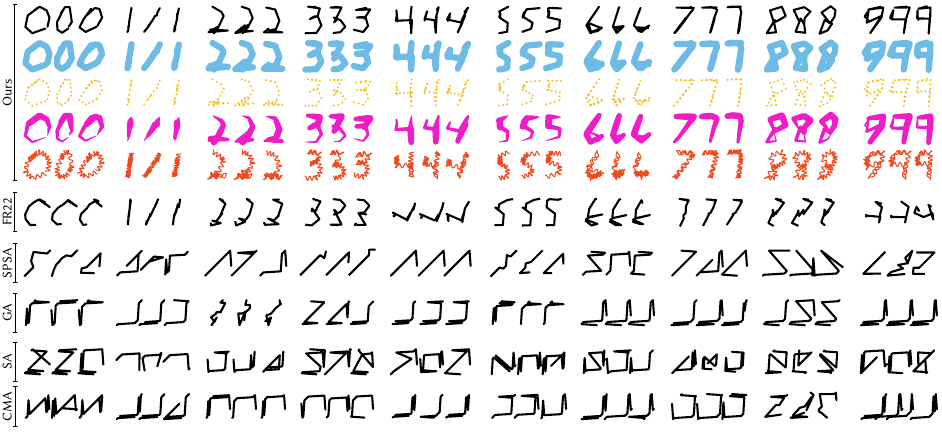}{We sample the latent space of our trained \ac{VAE} and show a variety of style transformations (rows), enabled by the spline formulation, on three digits per MNIST class. The first row displays the output of the spline renderer on which the surrogate operates.}
\task{Mesh} Next, we optimize the 3D position of \numvertices vertices of a triangle mesh, as in \citet{Nicolet2021Large}, to match a reference (\refFig{mesh}). 
This problem is already much harder, as the vertices are interlinked and the loss landscape exhibits discontinuities due to the rasterization process. 
On this task, \almostMethod{\ac{GA}} and \almostMethod{SPSA} fail to converge to the correct result, while \almostMethod{\ac{SA}} does not move from the initial configuration. 
In contrast, \almostMethod{CMA}, \almostMethod{FR22} and \method{Ours} find the correct solution, with \almostMethod{FR22} and \almostMethod{Ours} achieving the lowest final optimization errors (0.0018 and 0.0013, respectively). 
For completeness, we also show the differentiable rendering solution proposed by \cite{Nicolet2021Large}.

\mymath{\causticdim}{1,024}
\task{Caustic} In caustic optimization, a classic task from inverse rendering \cite{papas2011goal, schwartzburg2014high, kiser2013architectural}, the goal is to optimize an initial height-field such that it refracts incoming light into a caustic that resembles a provided reference image. 
While previous systems have gone to great effort to accurately capture the intricate behaviour of the refracted light, we use a simple rasterization-based renderer inspired by \cite{wyman2006interactive} (for details see \refSuppTaskDescr). 
Our heightfield is parameterized by a cubic B-Spline with resolution \causticdim. 
This example is interesting as the optimization variables have a highly non-local influence on the observed image pixels.
We show the resulting optimization outcomes in \refFig{caustic} and observe that all traditional gradient-free optimizers fail to correctly recover the target image. 
Again, \almostMethod{SPSA} and \almostMethod{FR22} achieve a caustic that roughly resembles the reference, while our method achieves a plausible outcome. 

\task{Spline Generation} Finally, we use our method to train a \emph{generative model} on a \emph{non-differentiable} task. 
Here, we use our surrogate gradients to train a \ac{VAE} \cite{kingma2013auto} that encodes digits from the MNIST \cite{lecun1998mnist} dataset and outputs the support points of a spline curve, which are then rendered with a non-differentiable spline renderer. 
As in all tasks, we operate in \emph{image space} only, so we cannot simply backpropagate through the spline renderer, but must query our surrogate for gradients.

This again is a very high-dimensional, non-local and interlinked problem, as all optimization variables (the \ac{NN} weights) directly influence the spline's final support points.
For details on the training and model architecture, please see \refSuppTaskDescr\!. 
In \refFig{MNISTStyle_large}, we sample the latent space of the model after training.
We render the generated spline in different styles, which can easily be applied in post-processing due to the control-point formulation.
As is evident from the figure, our method is the only approach that achieves an output that resembles actual digits across all numbers, with \almostMethod{FR22} achieving satisfactory results on simple cases (1, 3, 5, 7), and the other derivative-free optimizers failing completely. 
To our knowledge, this is the first generative model that is trained on a non-differentiable task, which again highlights the generality of our proposed approach, \name, and gives rise to an exciting avenue of future research.

\mysubsection{Gradient Variance Analysis}{gradientvariance}
While our method works well in all the previous tasks, its benefits are most pronounced when the loss landscape exhibits stochasticity or noise, e.g., in the \task{MLP} and \task{Splines} tasks. 
We hypothesize that this can be explained by the centerpiece of our approach, the \emph{neural proxy}: in contrast to \almostMethod{FR22} and \almostMethod{SPSA}, \name uses a neural network as proxy function, whose state acts as hysteresis and endows our method with a certain inertia, limiting the estimated loss landscape's spatiotemporal change by the network's adaptability. 
This behaviour is further reinforced by the spectral bias of neural networks \cite{rahaman2019spectral, tancik2020fourier}, which has been shown to encourage the learning of low-frequent, Lipschitz-continuous functions over those characterized by rapid changes.

\almostMethod{FR22} and \almostMethod{SPSA}, in contrast, re-build a (linear) gradient estimate during every iteration of the optimization, effectively ignoring information about the loss landscape from previous iterations. 
As this gradient estimate is a stochastic approximation, it will exhibit noise and variance, which highlights the main difference between the discussed (gradient-based) approaches: while \almostMethod{SPSA} and \almostMethod{FR22} estimate the \emph{parameter gradient} $\partial_\parameters$ (subject to variance), \name estimates the surrogate gradient $\partial_\surrogateParameters$, but \emph{analytically} computes the parameter gradient $\partial_\parameters$. 
This allows us to move the higher-variance estimate into the neural network's parameter update, where the estimate's noise is smoothed by the aforementioned hysteresis. 

To analyze this behaviour, we plot the variance of the gradient-magnitude over the course of the optimization in \refFig{variance} for all three\footnote{We exclude finite differences from this comparison due to its intractable per-iteration cost in higher dimensions.} gradient-based optimizers \almostMethod{SPSA}, \almostMethod{FR22} and \method{Ours}.
The vastly different scales on the y-axes of each subplot confirm our hypothesis: the variance in the gradient-magnitude of our method is consistently orders of magnitude lower than that of the other approaches. 
While this does not allow reasoning about the \emph{correctness} of the derived gradients, it explains why our approach outperforms the competitors in the provided examples.

\myfigure{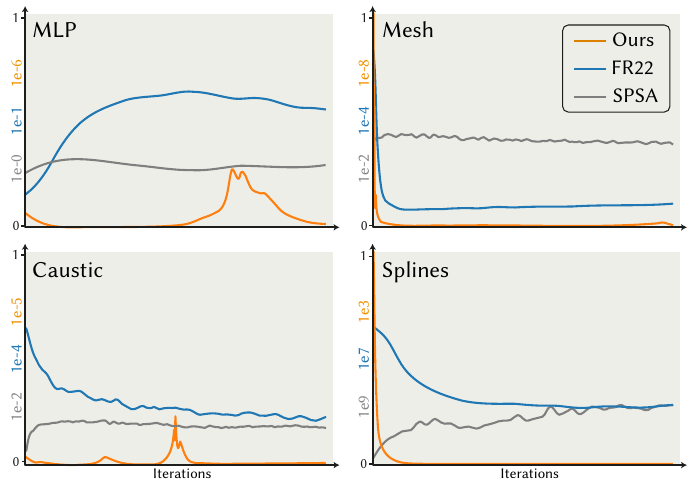}{We plot the variance (smoothed for ease of visualization) of the gradient magnitude over the course of the optimization. Note that the scales are vastly different, as denoted on the left axis. Our method consistently produces gradients with lower-variance magnitude, which we attribute to the neural proxy's state and smoothness, resulting in less gradient noise.}

\mysubsection{Comparison to specific solutions}{moreexperiments}

There exist many specialized solutions that enable gradient computation in graphics, and a full study of \emph{all} is beyond the scope of this work.
We compare qualitatively to two of these methods, rendering (\refFig{comparison_mitsuba}) and procedural modelling (\refFig{comparison_ours_hu22}), where the common theme is that our neural surrogates are capable of optimizing their specific problems as well, and sometimes even go beyond.
In \refFig{comparison_mitsuba}, top row, we show that we can optimize a material's \ac{IOR}, a feature for which backpropagation through detached sampling has not yet been implemented in Mitsuba. 
As our method only needs a forward-model, we can simply combine Mitsuba's forward path tracer with our surrogate gradients and thus are able to optimize the \ac{IOR} as well. 
In \refFig{comparison_ours_hu22}, we optimize a node graph towards the target patterns, using a mixture of VGG and MSE loss, which nicely shows our surrogate's flexibility w.r.t. to the objective \objective. 
Moreover, our method also works in extreme parameter ranges, as is evident from the bottom row in \refFig{comparison_ours_hu22}, where the pre-trained proxy from \citet{hu2022node} breaks due to the parameter value being out of the range it encountered during training. 
In summary, although our method might sometimes come at the expense of higher compute or variance (\eg compared to Mitsuba in \refFig{comparison_mitsuba}, see the convergence plots to the right), the strength of our approach lies in its generality, i.e., in that it can be applied to arbitrary forward graphic models, and that it successfully optimizes high-dimensional, interlinked problems. 

\myfigure{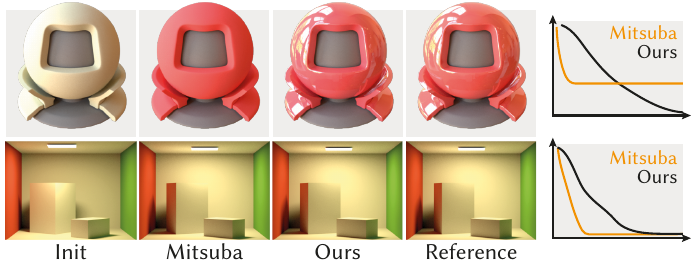}{Comparison with Mitsuba 3 \cite{jakobmitsuba} on the \task{BRDF} and \task{Cornell-Box} tasks, top and bottom row, respectively. Note that the incorrect IOR in the top row is due to Mitsuba not yet implementing this feature instead of failing during optimization (see \refSec{moreexperiments}).}
\myfigure{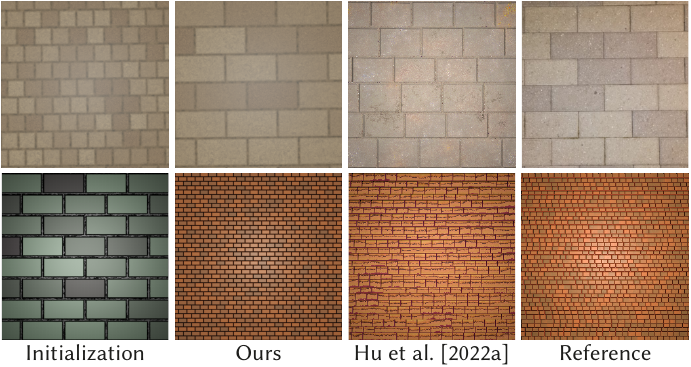}{Comparison between Ours and \citet{hu2022node} (result and reference taken from their publication). In the lower row, their pre-trained brick generator fails, as the parameter lies outside the training domain.}

\changed{\mysubsection{Limitations and Failure Cases}{Limitations}}
Our method inherits the limitations of gradient descent, namely that it can get stuck in local minima (although we do our best to avoid this via the smoothing convolution), move slowly in regions of shallow slope (see the experiments on the Rosenbrock function in \refSuppFigRosenbrock) and that it introduces additional hyperparameters (\refSuppHyperparams). 
Additionally, our method ``wastes'' samples during an initial warm-up phase, in which the (initially random) network weights first adapt to the loss landscape. 
Moreover, while our derived gradients have lower variance than competing approaches (\refFig{variance}), they have higher variance than analytical gradients \changed{and therefore typically under-perform relative to problem-specific methods, where they are available} (see the convergence plots in \refFig{comparison_mitsuba}). 
Finally, on lower-dimensional discrete problems, it can potentially be faster to simply brute-force the solution by trying all possible combinations, akin to what genetic algorithms would do with a high-enough sample budget. 
However, this quickly becomes infeasible as dimensionality increases. 

In addition, we show a failure case in \refFig{failurecase}. 
The task is inspired by PSDR-Room \cite{yan2023psdr,nguyen20123d} and the optimizer is asked to replicate a scene layout and materials from a single photograph.
For each piece of furniture or shrubbery, the optimizer can make a discrete choice from 10 objects (left column in \refFig{failurecase}) and additionally adapt their continuous 3D position in the scene and the wall's color (right column in \refFig{failurecase}).
We optimize MSE in the single-stage, single-resolution setting. 
While successful in the type-only setting, \name fails to correctly optimize both type and position in the right column of \refFig{failurecase}.
\myfigure{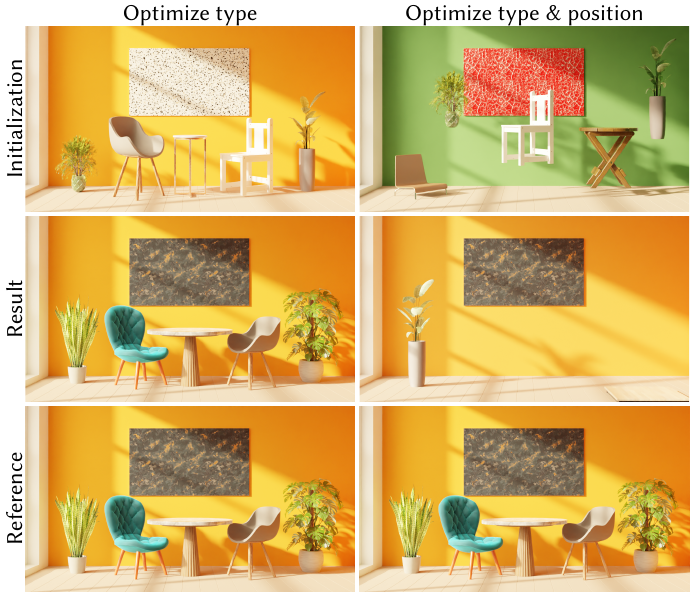}{A limitation of our method: when the loss landscape is too complex (right column: mixture of plateaus and discrete spaces), the proxy cannot encode it accurately and the optimization stalls in a local minimum.}
We assume this is because the optimizer must cycle through a number of objects before encountering the correct one, while simultaneously working in discrete and continuous space, and moreover dealing with plateaus in image-space which stem from the objects not overlapping their reference counterparts. 
Not even the smoothing operation can make this task easier, as the optimizer has the opportunity to reduce the image error by simply pushing the objects out-of-frame (as is happening here) and then falls into a local minimum from which it cannot recover. 
We conclude that more research is needed in this direction, e.g., through self-adapting proxy configurations or advanced hybrid approaches. 

\vspace{0.3cm}
\mysection{Conclusion}{Conclusion}
We have proposed \name, a method for practical computation of surrogate gradients in non-differentiable black-box pipelines, as are found across many graphics domains, ranging from rendering over modelling to animation.
Our key ideas are the smoothing of the loss landscape, a local approximation thereof by a \ac{NN}, and a low-variance estimator based on sparse, local sampling.
We have favourably compared to several ablations and published alternatives and shown results for a wide variety of tasks. 
Additionally, our neural surrogate allows us to transform the noisy gradient estimate into an update on the network's parameters, where the noise is smoothed by the network's hysteresis.
We therefore can show that our surrogate gradients scale to high dimensionality, where traditional gradient-free optimization algorithms often do not converge. 

In future work, we plan to further explore the interplay of the inner surrogate loss and the outer optimizer and to find ways to automatically determine the required surrogate network's complexity.  
Moreover, it would be interesting to leverage the fact that our surrogate provides a continuous model of the cost landscape, for instance by lookahead-training or approximate second-order methods. 

Most of the things that enable our approach are known in the optimization community that routinely uses proxies and surrogates.
Yet, these ideas are rarely used in graphics, where specific solutions were developed and rarely compared against what the optimization literature offers.
Our work combines graphic-specific features (e.g., MC-estimating the gradient, sampling the objective through simply rendering) and graphics-inspired improvements (such as variance reduction through importance sampling) to match requirements of graphics with general optimization.

\begin{acks}
This work was supported by \grantsponsor{MRL}{Meta Reality Labs}{}, Grant Nr.\ \grantnum{MRL}{5034015}. 
Michael is further supported by a bursary from the Rabin Ezra scholarship trust. 
\end{acks}

\bibliographystyle{ACM-Reference-Format}
\bibliography{paper}

\mycfigure{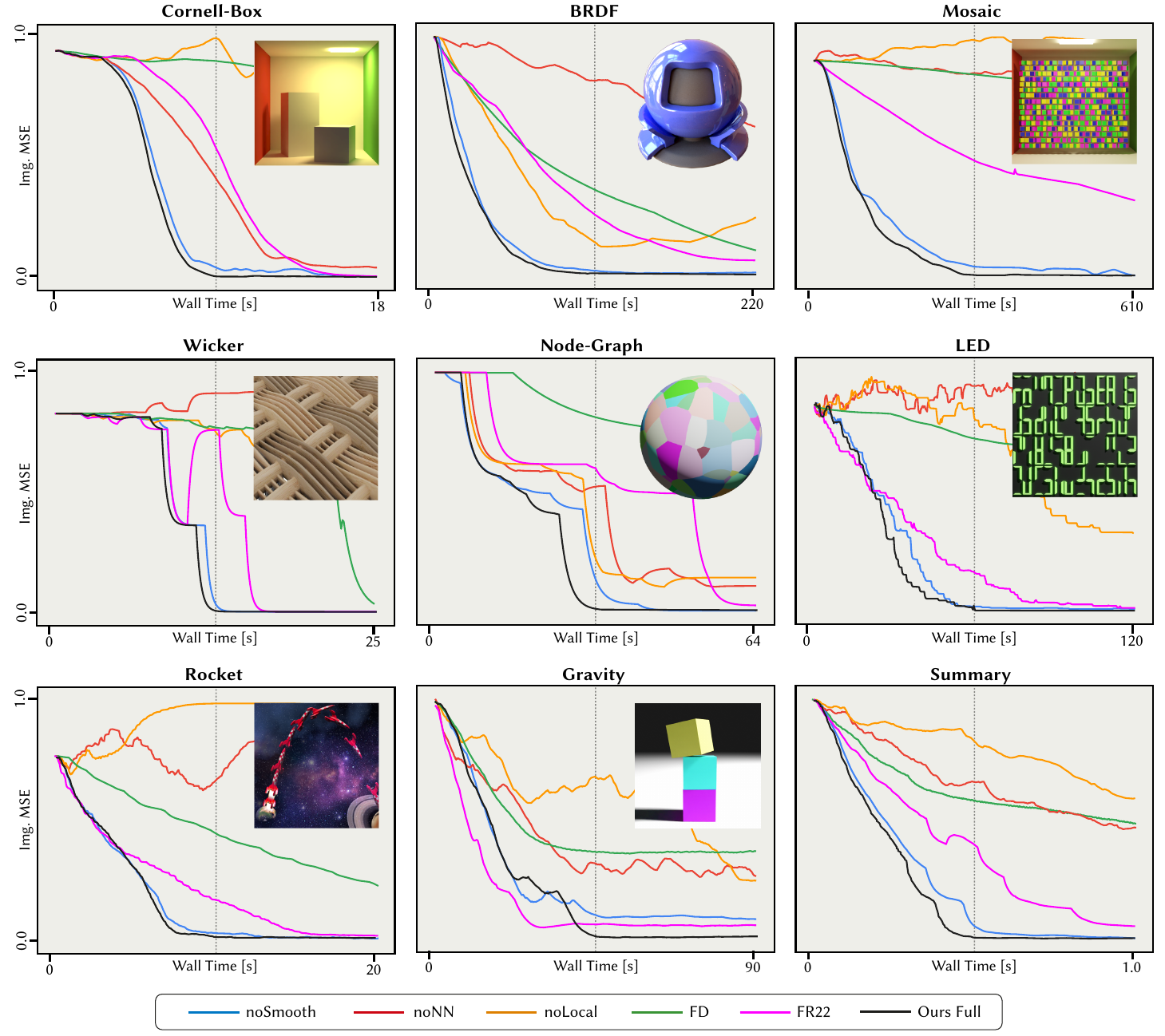}{We show convergence plots of all methods (wall-clock time in seconds, method colors are consistent with \refTab{overview}) for all tasks, ranging from differentiable rendering (top row) over procedural modelling (middle row) to animation and simulation (bottom row).  
For all experiments, we let our method run until convergence (the dashed vertical line in each subfigure) and then allocate \emph{twice as much} time for the other methods to converge. All results are reported across an ensemble of 10 independent runs for all methods. For convenience, we show a summary across all tasks in the right bottom subfigure (mean across all methods, normalized and resampled). For task-specific visualizations, please see \refFig{results_qualitative}.}

\mycfigure{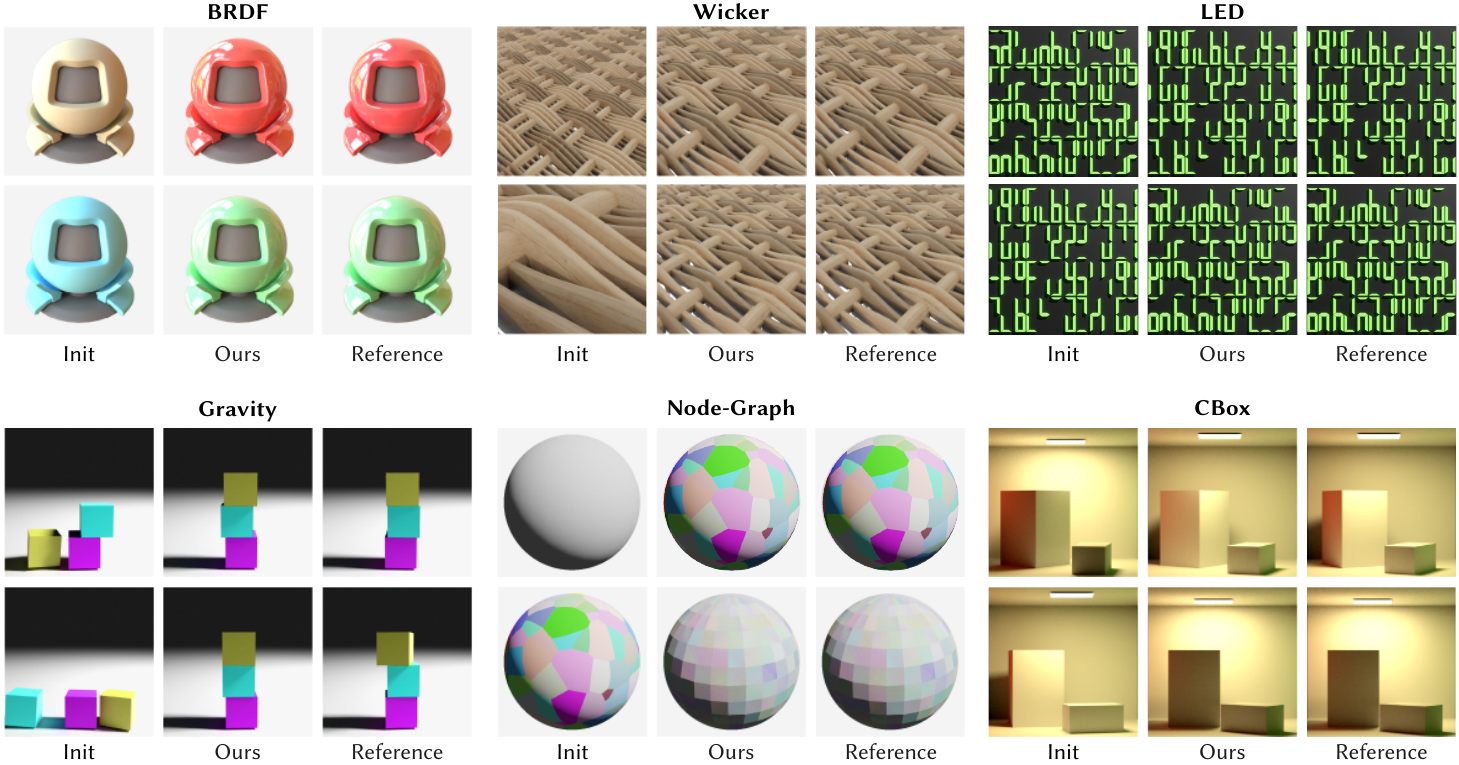}{Visualizations of instances of our tasks. We visualize the initial configuration in the left column of each subfigure, and our outcome and the reference in the middle and right column, respectively. For our method's convergence behaviour on these tasks, please see \refFig{results}.}

\end{document}


\title{Zero Grads: Supplemental Materials}

\author{Michael Fischer}
\affiliation{%
	\institution{University College London}
	\country{United Kingdom}
}
\email{m.fischer@cs.ucl.ac.uk}

\author{Tobias Ritschel}
\affiliation{%
	\institution{University College London}
	\country{United Kingdom}
}
\email{t.ritschel@ucl.ac.uk}

\maketitle

\acresetall

\mymath{\adjoint}a
\mymath{\lossOffset}{\rho}
\mymath{\learningRate}{\alpha}
\mymath{\locality}{\lambda}
\mymath{\mcmcSample}{y}
\mymath{\mcmcSampleGradient}{\mathbf z}
\mymath{\mcmcState}{\bar}
\mymath{\objective}f
\mymath{\offset}{\tau}
\mymath{\parameters}{\theta}
\mymath{\parameterDimension}n
\mymath{\parameterSpace}{\Theta}
\mymath{\sampleCount}{N}
\mymath{\smoothObjective}g
\mymath{\smoothingKernel}{\kappa}
\mymath{\spreadinner}{\sigma_i}
\mymath{\spreadouter}{\sigma_o}
\mymath{\surrogate}h
\mymath{\surrogateLoss}l
\mymath{\surrogateParameters}{\phi}
\mymath{\surrogateParameterSpace}{\Phi}
\mymath{\surrogateParameterDimension}m
\mymath{\potentialMatrix}{\textsf{M}}
\mymath{\inputvector}{\textrm{x}}

\mymath{\nsplinepts}{10}
\mymath{\numtexels}{196,608}
\mymath{\texturedim}{256}
\mymath{\numvertices}{2,562}
\mymath{\meshdim}{7,686}
\mymath{\numweights}{35,152}
\mymath{\numsplineweights}{8,764}

\mymath{\taskdim}{n_{\textrm{dim}}}
\mymath{\numcaustic}{16,384}
\mymath{\causticdim}{1,024}

This supplementary contains additional information on our surrogate implementation and hyperparameters (\refSec{Implementationdetails}), rendering setups (\refSec{Rendering}), and detailed descriptions of the tasks we solve  (\refSec{TaskDescriptions}). 

\mysection{Implementation Details}{Implementationdetails}

We implement all our experiments in PyTorch \cite{paszke2017automatic}.
The proxy powering our surrogate is implemented as a \ac{MLP} and activated by a leaky ReLU. 
We randomly initialize our Neural Proxy for each optimization run (via the standard PyTorch initialization, for the quadratic proxy, we choose the identity matrix) 
and optimize its weights alongside the parameter with a separate Adam optimizer.
We perform three update steps on the surrogate parameters \surrogateParameters per optimization iteration in order to improve the surrogate's fit to the sampled data.
This is simple autodiff-driven \ac{GD} and hence very fast. 
Note that no new data is sampled between these update steps, they merely serve to improve the surrogate fit and do not increase the required computational budget. 
For all gradient updates, we use the Adam optimizer with standard parameters 
and learning rates as specified in \refTab{hyperparameters}.
We additionally experimented with different sampling patterns and found both both low-discrepancy (Sobol) and antithetic samples and found both to improve performance, and adapt antithetic samples for simplicity. 
We normalize the network's inputs to [0,1]. 
For the lower-dimensional tasks (\taskdim < 50), it suffices to use 3 hidden layers with 64 neurons each, whereas for the higher-dimensional tasks (below the horizontal line in \refTab{hyperparameters}), we found that we needed to increase the surrogate's capacity to 8 layers \`a 128 neurons and additionally use positional encoding to increase the frequencies that the network can encode.

\mysubsection{Hyperparameters}{Hyperparameters}
Our method comes with two hyperparameters: the number of samples \sampleCount we use to estimate our surrogate's gradient with (cf. Alg.2 in the main text), and the spread of the locality kernel \locality, which will influence how far these samples are spaced out around the current parameter \parameters.

\mycfigure{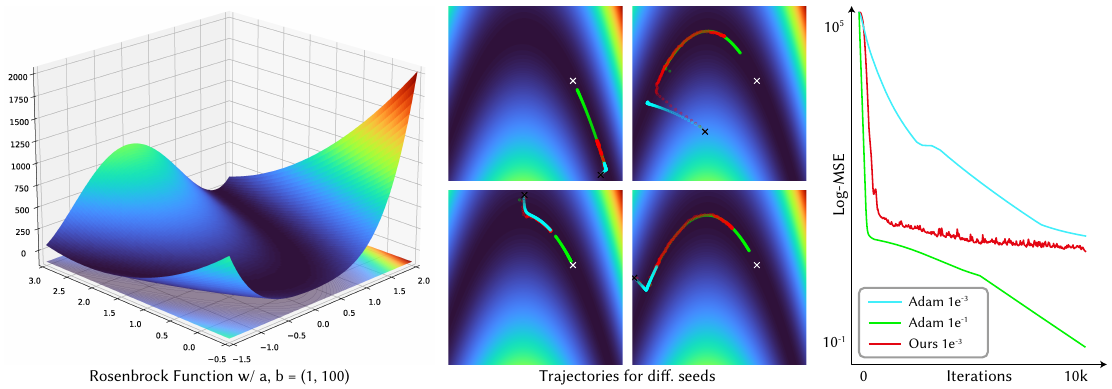}{We evaluate our method on the Rosenbrock function against gradient descent with analytical gradients and Adam with equal learning rate, sample count and iterations. Similar to Adam, our method struggles to make progress in the valleys of low slope, a common limitation of gradient-based techniques. Adam, with a higher learning rate, converges faster than our method. The convergence plots in the right subfigure are median values over an ensemble of 10 independent runs and seeds.}

\begin{table}[h!]
    \setlength{\tabcolsep}{3.5pt}
    \centering
    \caption{Our hyperparameters \spreadouter and \sampleCount, as well as the experiment settings for the different tasks, sorted by dimensionality in ascending order. MPL is short for matplotlib.}
    \begin{tabular}{lrrrrrr}
        & \multicolumn{1}{r}{\spreadouter} 
        & \multicolumn{1}{r}{\sampleCount} 
        & \multicolumn{1}{r}{\taskdim} 
        & \multicolumn{1}{r}{LR \parameters}
        & \multicolumn{1}{r}{LR \surrogateParameters} 
        & \multicolumn{1}{r}{Renderer} \\ 
        \toprule 
        \task{Wicker} & 0.33 & 2 & 3 &$1\!\times\!10^{-3}$ & $1\!\times\!10^{-3}$ & Blender \\
        \task{BRDF} & 0.33 & 2 & 4 &$1\!\times\!10^{-3}$ & $1\!\times\!10^{-3}$ & Mitsuba\\
        \task{CBox} & 0.10 & 2 & 4 & $5\!\times\!10^{-4}$& $1\!\times\!10^{-3}$ & Mitsuba\\
        \task{Gravity} & 0.20 & 2 & 5 &$1\!\times\!10^{-3}$ & $1\!\times\!10^{-3}$ & Blender \\     
        \task{Rocket} & 0.33 & 2 & 10 & $1\!\times\!10^{-3}$& $5\!\times\!10^{-4}$ & MPL \\
        \task{NodeGr.} & 0.20 & 2 & 24 &$1\!\times\!10^{-3}$ & $1\!\times\!10^{-3}$ & Blender \\
        \task{LED} & 0.33  & 2 & 336  &$1\!\times\!10^{-3}$ & $1\!\times\!10^{-3}$ & Blender \\
        \midrule
        \task{Mosaic} & 0.025 & 16 & 320 & $5\!\times\!10^{-4}$& $1\!\times\!10^{-3}$ & Blender\\
        \task{Caustic} & 0.013 & 20 & \causticdim & $2\!\times\!10^{-4}$& $1\!\times\!10^{-4}$ & PyTorch \\   
        \task{Mesh} & 0.025 & 20 & \meshdim & $2\!\times\!10^{-3}$& $1\!\times\!10^{-4}$ & NVDiff. \\     
        \task{Spline Gen.} & 0.025 & 20 & \numsplineweights &$1\!\times\!10^{-5}$ & $1\!\times\!10^{-4}$ & MPL \\     
        \task{MLP} & 0.025 & 20 & \numweights &$1\!\times\!10^{-4}$ & $1\!\times\!10^{-4}$ & MPL \\     
        \task{Texture} & 0.025 & 20 & \numtexels &$1\!\times\!10^{-5}$ & $1\!\times\!10^{-4}$ & MPL \\     
        \bottomrule 
    \end{tabular}
    \label{tab:hyperparameters}
\end{table}

We specify the number of samples \sampleCount we use for estimating the surrogate's gradients in \refTab{hyperparameters}. 
For the lower-dimensional tasks, it suffices to use $\sampleCount=2$, whereas for the higher-dimensional tasks, the noise and higher variance from this rough gradient estimate impede convergence and thus require higher sample counts. 
We would like to emphasize that those are still far lower than what competing methods use, e.g., 2\taskdim for \ac{FD} or $m \,\times\, \taskdim, m \gg 2$, for \ac{DGS} \cite{zhang2020scalable}. 
Our method also benefits from more samples in the lower-dimensional regime, but these come at the cost of increased compute, which is why we tried to achieve a minimal number to keep the overhead low. 

\myfigure{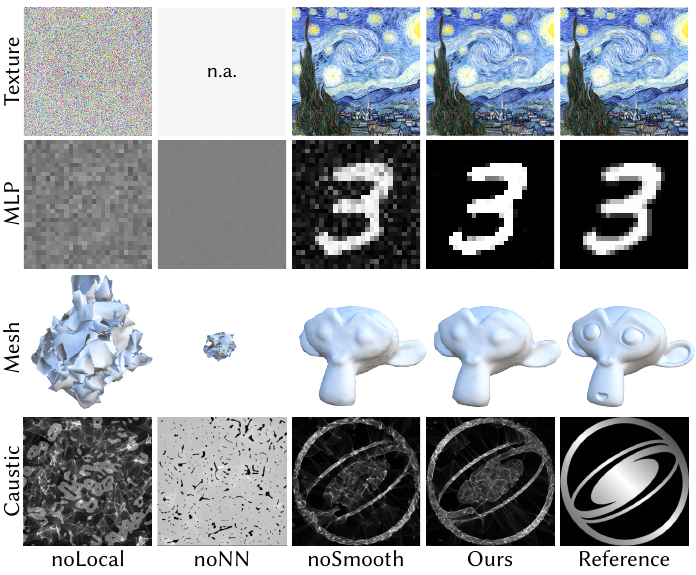}{We show the results of our ablated methods from the main manuscript (\refMainMethods) on the higher-dimensional tasks. Similar to CMA, the result for the quadratic proxy (noNN) could not be run due to the quadratic memory complexity.}

\begin{wrapfigure}[8]{r}{0.5\columnwidth}%
\vspace{-.2cm}%
\centering\includegraphics*[width=\linewidth]{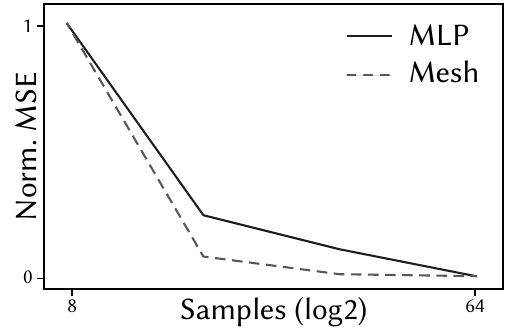}%
\vspace{-.3cm}%
\caption{Final error vs. samplecount \sampleCount.}
\label{fig:samplecount}
\end{wrapfigure}

We show a comparison of different sample counts on the \task{Mesh} and \task{MLP} tasks in \refFig{samplecount} and detail the remaining hyperparameters and experiment settings in \refTab{hyperparameters}, where \spreadouter denotes the spread of the locality kernel \locality. 
As a general rule of thumb, we recommend setting the initial \spreadouter to 0.33 on normalized domains and 
fine\-tune from there, if necessary.
For higher-dim\-en\-sional, interlinked problems, we have found a more fine-granular sampling to be nec\-es\-sary and use $\spreadouter=0.025$.
We use 15\% of the locality spread as the spread of the smoothing kernel \smoothingKernel.

\mysection{Tasks}{tasks}
This section provides information on the task setup, problems, goals and rendering architectures used. 

\mysubsection{Rendering}{Rendering}
To render the images for the tasks Mosaic, Wicker, LED, NodeGraph and Gravity, we interface our method with Blender via an efficient socket-based local TCP network, which enables us to make use of Blender's rendering engines and the embedded physics solver. 
All images were set to render noise-free under either Eevee or Cycles, with 16 to 128 samples and denoising activated.
For the tasks BRDF and Cornell-Box, and the comparisons with Mitsuba, we use Mitsuba 3 \cite{jakobmitsuba} with the path-replay backpropagation integrator at 16spp. 
For the Mesh task, we use NVDiffRast \cite{laine2020modular} with standard hyperparameters. 
For the remaining tasks Rocket, Spline Generation and Texture, we use a custom matplotlib-based renderer \cite{hunter2007}. 
Note that none of this interfacing is necessary for our method to work, but pure convenience for rapid prototyping and reducing I/O times from and to disk. 
Most importantly, we do not propagate any gradient information through the rendering process, even if this were possible, e.g., when using a differentiable renderer. 
One could alternatively render an image, save it to disk and manually load it and perform a gradient update step, which would yield the same results, but be arguably less convenient. 

\mysubsection{Task Descriptions}{TaskDescriptions}

\mysubsubsection{Higher Dimensions}{HigherDimensions}
While some of our higher-dimensional example tasks could in theory also be solved via established, specialized methods (e.g., \cite{jakobmitsuba, Nicolet2021Large, holl2020learning}), they show that our method scales well to higher dimensional problems and reinforce our argument of general applicability. 
All comparisons to the following optimization algorithms are performed under the same budget of function evaluations. 

For the comparisons with \acp{GA}, we use the publicly available Python package pygad \cite{gad2021pygad}.
For \ac{SA} \cite{xiang2013generalized}, 
we use the \texttt{scipy} library \cite{virtanen2020scipy}.
For \ac{SPSA} \cite{spall1992multivariate}, we use the publicly available \texttt{spsa} package \cite{spsapackage}. 
Note that, while we use standard hyperparameters for the other packages, we here adapted the \ac{SPSA} perturbation radius to the sampling radius used by our method in order to enable a fair comparison (the default value of 2.0 is too large for many of our problems, e.g., for the delicate task of network training). 

\task{Texture} For the \task{Texture} task, we use our method to optimize the \texturedim pixels of an image texture, leading to a $\texturedim \!\times\! \texturedim \!\times\! 3 = \numtexels$ optimization problem. 
We randomly initialize the texels from $\mathcal{N}(0.5, 0.05)$, i.e., they are drawn from a Normal distribution with mean 0.5, corresponding to a grey value.
As is common, we additionally employ a whitening transform during optimization \cite{NimierDavidVicini2019Mitsuba2}.

\task{MLP} This task is an extension of the texture task to address the concern that optimization variables are not sufficiently interlinked with each other.
To this end, we train a \ac{MLP} to replicate randomly sampled digits from the MNIST \cite{lecun1998mnist} dataset. 
The MLP has two ReLU-activated hidden layers of 32 neurons and a final layer with 784 neurons that is activated by a Sigmoid, leading to a total of \numweights network weights and hence to a \numweights-dimensional optimization problem. 
The weights are initialized via the standard formula $\mathcal{U}(-k, k)$, where $k$ is the reciprocal of the layer's input features \cite{paszke2017automatic}. 

\task{Caustic} For this task, we take inspiration from \citet{wyman2006interactive} and write a fast, rasterization-based caustic renderer.

\mywfigure{rocket_example}{0.44}{An illustration why differentiating an ODE solver $\mathcal{R}(\theta)$ w.r.t. time is not trivially differentiable: moving the event-time $\theta$ of the blue signal within the yellow interval will not affect the observed outcome, as the solver operates on the discretized version \textcolor{red}{$\widehat{\mathcal{R}}$} only and will continue to observe ``on'' and ``off'' at timesteps $i$ and $i+1$, respectively.}
The idea is that a parallel bundle of rays from a faraway directional light source hits a parameterized refractive surface (our heightfield, usually modeled as a glass slab \cite{NimierDavidVicini2019Mitsuba2, papas2011goal, schwartzburg2014high}), and gets refracted according to Snell's law (we use an index of refraction of 1.33). 
The refracted rays then hit a receiver plane, where we record, for each pixel, the number of received rays, resulting in an approximate caustic. 
We use an equal ray- and receiver resolution of 512p.
The relation between the optimization variables (the heightfield, in our case parameterized as a cubic B-Spline of resolution $32^2$, randomly initialized) and the final output in this task is highly non-linear, as a change in the heightfield has the potential to affect various pixels across the entire receiver plane. 
Moreover, the task is not trivially differentiable, as the conversion of the (continuous) hitpoint on the receiver plane to discrete pixel coordinates in the image grid is a discontinuous operation. 

\task{Mesh} For the \task{Mesh} task, we optimize the vertices of a triangle mesh such that the renderings of the mesh match those of a reference shape. 
Our source mesh has \numvertices vertices whose 3D positions we optimize, leading to a highly interlinked \meshdim-dimensional problem.
We follow the approach in \cite{Nicolet2021Large} and use their smooth formulation, the AdamUniform optimizer and the Laplacian regularization, thereby nicely showing that our surrogate successfully learns to replicate the regularized loss landscape. 
For fairness, all competitors operate in this parametrization. 
Following \cite{Nicolet2021Large}, the source shape is initialized as a tessellated sphere and rendered from 13 different viewpoints under environment illumination using NVDiffRast \cite{laine2020modular} -- however, without backpropagating their gradient information; all gradients employed in the optimization are produced by our surrogate. 

\task{Spline Generation} For the \task{Spline Generation} task, we train a generative model, a \ac{VAE}\cite{kingma2013auto}, to replicate digits from the MNIST dataset in a spline representation. 
Our \ac{VAE} consists of an encoder-\ac{MLP} with roughly 40k neurons, and a decoder-\ac{MLP} with \numsplineweights neurons. 
To stabilize training, we use a pre-trained encoder that serves as feature extractor and projects the MNIST images into the latent space, from where we learn a generative decoder that predicts the horizontal and vertical translation of \nsplinepts spline support points (initialized diagonally across the image plane). 
Subsequently, we fit a spline through these predicted support points with a (matplotlib-based) non-differentiable renderer and learn our surrogate on the reconstructed splines' image-space \ac{MSE}, regularized by the \ac{VAE}'s \ac{KLD} (weighting factor $0.1$). 
Descending along the surrogate gradients then produces the weights for a generative decoder that can be sampled to generate new MNIST digits. 
Again, we initialize all stateful components with the standard formula $\mathcal{U}(-k, k)$, where $k$ is the reciprocal of a layer's input features \cite{paszke2017automatic}. 

\bibliographystyle{ACM-Reference-Format}
\bibliography{paper}